\documentclass[sigconf,nonacm]{acmart}
\usepackage{booktabs, array, multirow}

\usepackage{amssymb}
\AtBeginDocument{%
  }

\setcopyright{none}

\begin{document}

\title{Traffic-CBM: A Structurally Interpretable Multimodal Framework for Encrypted Traffic Classification}

%

\author{Honglei Jin}
\email{velix.lab@gmail.com}
\affiliation{%
  \institution{HKUST (GZ)}
  \city{Guangzhou}
  \country{China}
}

\author{Wenshuo Chen}
\email{wchen179@connect.hkust-gz.edu.cn}
\affiliation{%
  \institution{HKUST (GZ)}
  \city{Guangzhou}
  \country{China}
}

\author{Shaofeng Liang}
\email{shawnsfliang@hkust-gz.edu.cn}
\affiliation{%
  \institution{HKUST (GZ)}
  \city{Guangzhou}
  \country{China}
}

\author{Haozhe Jia}
\email{haozhezzz@outlook.com}
\affiliation{%
  \institution{HKUST (GZ)}
  \city{Guangzhou}
  \country{China}
}

\author{Runwei Guan}
\email{runwayrwguan@hkust-gz.edu.cn}
\affiliation{%
  \institution{HKUST (GZ)}
  \city{Guangzhou}
  \country{China}
}

\author{Menshuo Zhao}
\email{zhaomengshuoexo@gmail.com}
\affiliation{%
  \institution{Shandong University}
  \city{Qingdao}
  \country{China}
}

\author{Shuxu Jin}
\email{shuxujin0315@163.com}
\affiliation{%
  \institution{Shandong University}
  \city{Qingdao}
  \country{China}
}

\author{Songning Lai}
\email{lais0328eee@gmail.com}
\affiliation{%
  \institution{HKUST (GZ)}
  \city{Guangzhou}
  \country{China}
}

\author{Yutao Yue}
\authornote{Correspondence to Yutao Yue <yutaoyue@hkust-gz.edu.cn>}
\email{yutaoyue@hkust-gz.edu.cn}
\affiliation{%
  \institution{HKUST (GZ)}
  \city{Guangzhou}
  \country{China}
}

\settopmatter{authorsperrow=3}

\renewcommand{\shortauthors}{Honglei Jin et al.}

\begin{abstract}
Encrypted traffic classification has achieved strong performance, but its decision process remains difficult to interpret. Existing methods usually combine flow statistics, packet sequences, and byte-level representations into opaque latent features, making it unclear which type of evidence actually drives the prediction. In this paper, we propose Traffic-CBM, a structurally interpretable multimodal framework for encrypted traffic classification. Instead of directly fusing heterogeneous traffic signals into a black-box representation, Traffic-CBM organizes them into a unified hierarchical concept space. These concepts are not manually annotated semantic attributes; rather, they are scalar evidence summaries constrained by predefined traffic evidence groups. More specifically, grouped flow statistics are mapped to statistical concepts, dedicated temporal encoders learn temporal concepts from disjoint feature subspaces, and byte-level evidence is further organized into packet-level and cross-packet concepts. This design turns heterogeneous traffic evidence into an explicit concept representation and makes different levels of traffic evidence easier to analyze. We evaluate Traffic-CBM on multiple encrypted traffic benchmarks. Results show that it achieves competitive and balanced classification performance while providing a clearer structural interpretation interface than conventional end-to-end fusion models. Further analyses suggest that the learned concept space is actively used in the prediction process and provides a clearer structural explanation of multimodal traffic evidence.
\end{abstract}

\keywords{Encrypted Traffic Classification, Multimodal Learning, Structural Interpretability, Concept-based Modeling, Interpretable Deep Learning}


\maketitle

\section{Introduction}

\begin{figure}[t]
    \centering
    \includegraphics[width=\columnwidth]{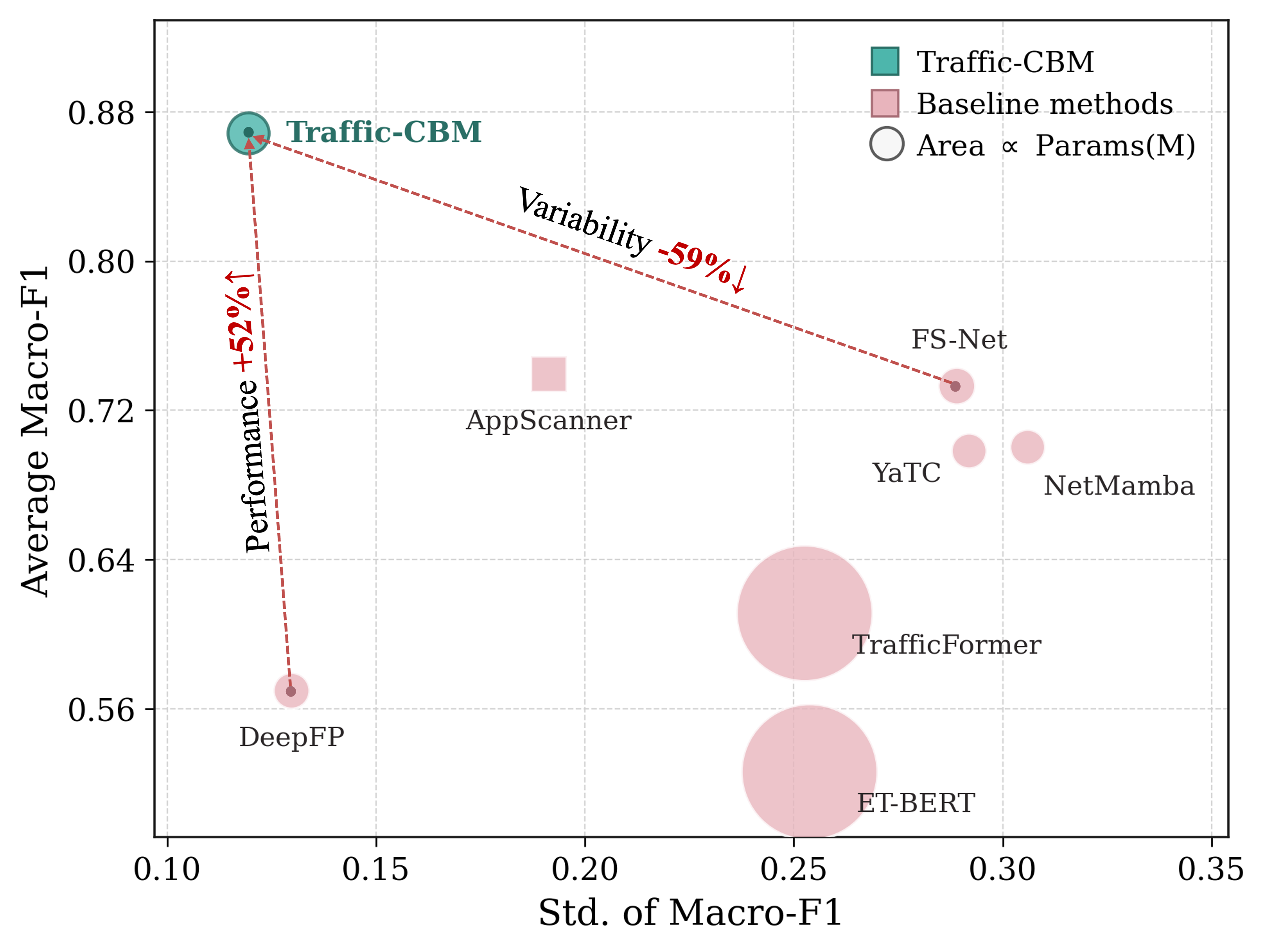}
    \vspace{-16pt}
    \caption{Overall performance versus cross-dataset stability across six datasets. The x-axis denotes the standard deviation of Macro-F1 across datasets (lower is better), and the y-axis denotes the average Macro-F1 (higher is better). Bubble area is proportional to parameter count for neural methods, while the non-neural baseline AppScanner is shown with a fixed marker size. Methods closer to the upper-left region achieve a more favorable overall trade-off. Traffic-CBM achieves a favorable balance between average performance, stability, and model size.}
    \label{fig:overall_balance}
    \vspace{-14pt}
\end{figure}

Encrypted traffic classification has become increasingly important for understanding modern networked applications, since a large portion of Internet services is now delivered under encryption~\cite{Dong2025Review,Sharma2025Survey}. In practical security scenarios, accurate predictions alone are often not sufficient; analysts also need to understand what traffic evidence supports a decision and whether the model is relying on stable behavioral patterns or spurious cues~\cite{nascita2021xai,Hermosilla2025XAIIDS}. Recent learning-based methods have substantially improved classification performance by exploiting heterogeneous traffic evidence, including flow statistics, packet sequences, and byte-level representations~\cite{fsnet2019,etbert2022,Dong2025Review}. Together, these signals form a multimodal traffic representation for classification~\cite{Dong2025Review}. However, despite these advances, interpretability in encrypted traffic analysis remains underdeveloped.

This limitation is especially challenging in encrypted settings. Although payload bytes may remain observable, their plaintext semantics are hidden by encryption, making it difficult to interpret model decisions directly and to determine how different traffic evidence contributes to prediction~\cite{nascita2021xai,Hermosilla2025XAIIDS}. In addition, encrypted traffic classifiers may suffer from dataset-specific biases and spurious correlations~\cite{sok2025}, which makes interpretable models particularly valuable for diagnosing model behavior and validating security decisions in high-stakes environments~\cite{nascita2021xai,Hermosilla2025XAIIDS}.

Most existing studies on explainability for encrypted or network traffic analysis rely on post-hoc interpretation, such as SHAP-, LIME-, or attribution-based analyses applied after training to identify influential input features~\cite{nascita2021xai,Nascita2023ExplainableTC,Suh2025SHAPTraffic}. While such methods can provide useful local clues, they do not explicitly constrain how heterogeneous evidence is organized inside the model, and thus offer limited structural insight into whether a model bases its decisions on stable high-level behavioral factors or on entangled low-level cues. In parallel, many end-to-end fusion models combine heterogeneous traffic modalities into a single predictive representation~\cite{etbert2022,yatc2023,netmamba2024,trafficformer2025}. Existing methods often rely on specific types of traffic evidence, and their performance can vary across datasets where the most informative signals differ. As a result, evidence from different abstraction levels is often mixed together, leaving its internal organization structurally unconstrained. Intrinsic concept-based modeling, which could help address these limitations, remains largely unexplored in encrypted traffic classification.

Rather than explaining a black-box model only after training, we argue that interpretability in encrypted traffic classification should be built into the modeling process itself. Specifically, heterogeneous traffic evidence should be organized explicitly during modeling, in the spirit of concept-based and concept-bottleneck modeling~\cite{cbm2020,cat2024}. In encrypted traffic classification, the relevant concepts are more naturally tied to observable traffic evidence groups, such as timing, length, TCP behavior, and byte-level regularities, rather than to manually named semantic attributes. Based on this view, we propose \textbf{Traffic-CBM}, a structurally interpretable multimodal framework for encrypted traffic classification.

The key idea of Traffic-CBM is to build prediction on a hierarchical concept space rather than on an entangled multimodal latent representation. Flow statistics are mapped into grouped statistical concepts, packet-level sequential features are processed by dedicated temporal encoders over disjoint feature subspaces, and byte-level evidence is explicitly decomposed into \emph{packet-level concepts} and \emph{cross-packet concepts}. This split is motivated by the observation that byte-level traffic evidence contains both packet-local patterns and cross-packet regularities, which are difficult to disentangle when modeled by a single shared byte representation. In this way, Traffic-CBM turns heterogeneous traffic evidence into an explicit concept representation and makes the roles of different evidence sources easier to analyze.

Traffic-CBM provides a structurally interpretable concept space and supports concept-level analysis through contribution, interaction, faithfulness, and sufficiency. In addition, we present a selected input-level validation case study for concept pathways that can be linked back to original traffic evidence. We evaluate Traffic-CBM on multiple encrypted traffic benchmarks and compare it with existing baselines as well as alternative concept modeling variants. As summarized in Figure~\ref{fig:overall_balance}, Traffic-CBM achieves a favorable balance between average performance, cross-dataset stability, and model size, while offering a clearer structural interpretation interface than conventional end-to-end fusion models.

In summary, this paper makes the following contributions:
\vspace{-2pt}
\begin{itemize}
    \item We formulate encrypted traffic classification as a structurally interpretable multimodal modeling problem and show why post-hoc attribution alone is insufficient for revealing how heterogeneous traffic evidence is internally organized.
    \item We propose Traffic-CBM, a structurally interpretable multimodal framework that organizes statistical, temporal, and byte-level traffic evidence into a unified hierarchical concept space for encrypted traffic classification, with byte-level evidence further structured into packet-level and cross-packet concept groups.
    \item We show across multiple encrypted traffic benchmarks that Traffic-CBM achieves competitive and balanced performance, and that its concept space supports concept-level analysis while providing a clearer structural interpretation interface for multimodal traffic evidence.
\end{itemize}
\section{Related Work}

\subsection{Encrypted Traffic Classification}

Encrypted traffic classification has advanced rapidly with the adoption of modern encryption protocols. Existing methods differ in traffic granularity, input representation, and feature extraction strategy, which in turn shape how statistical, sequential, and byte-level evidence is exploited~\cite{Dong2025Review,sok2025}. Early studies mainly relied on observable side-channel evidence, such as statistical patterns and packet-level sequences, for encrypted traffic classification and related encrypted traffic analysis tasks~\cite{appscanner2018,flowprint2020,hayes2016kfp,alnaami2016bind}. Subsequent deep models further reduced reliance on handcrafted features by learning directly from raw flow sequences, traffic patterns, or packet content, as exemplified by Deep Fingerprinting (DeepFP)~\cite{deepfp2018}, FS-Net~\cite{fsnet2019}, and Deep Packet~\cite{deeppacket2020}.

Recent methods can be roughly grouped by the type of traffic evidence they model. One direction focuses on direct traffic content, especially raw bytes or packet representations. Representative examples include Transformer-based models such as PERT, ET-BERT, and TrafficFormer~\cite{pert2020,etbert2022,trafficformer2025}, masked-autoencoder-style pre-training as in Flow-MAE~\cite{flowmae2023}, Mamba-based byte modeling as in NetMamba~\cite{netmamba2024}, and image-style byte modeling as in YaTC~\cite{yatc2023}. Another direction focuses more on transmission patterns, such as packet lengths, directions, timing information, and packet-level sequences, which remain informative even when payload semantics are unavailable~\cite{appscanner2018,flowprint2020,fsnet2019,tscrnn2021,graphdapp2021}. Recent studies on networked data and structured representation learning have begun to combine multiple evidence sources, domain constraints, or heterogeneous representation spaces within unified frameworks~\cite{mm4flow2025,jia2025pira,tian2025text2weight}. While these advances substantially improve accuracy, most existing classifiers remain primarily performance-driven, and heterogeneous traffic evidence is usually exploited within black-box predictive pipelines rather than explicitly organized into interpretable structures~\cite{Dong2025Review,sok2025}.

\subsection{Interpretable Machine Learning}

Interpretable machine learning broadly includes post-hoc explanation and intrinsic interpretable modeling. Post-hoc methods, such as LIME and SHAP, explain trained black-box models through feature importance or attribution analysis, but do not constrain how evidence is represented internally~\cite{lime2016,shap2017}. Intrinsic models instead build interpretability into the model structure itself, making it possible to analyze how heterogeneous evidence sources are organized before prediction.

Representative intrinsic models include generalized additive models and their extensions, such as GA2M and related neural variants like NAM, which explain predictions through additive main effects and pairwise interactions~\cite{Lou2013GA2M,agarwal2021nam}. More broadly, concept-based explainability uses higher-level semantic or structured factors as the basis of explanation, as exemplified by CBM and recent concept-based XAI surveys~\cite{cbm2020,poeta2025cxai}. Recent extensions further explore concept bottlenecks, additive concept-based prediction, and concept modeling in broader settings~\cite{oikarinen2023cbm,cat2024,tabcbm2023,hybridcbm2025}. However, existing label-free or tabular concept models, such as label-free CBM~\cite{oikarinen2023cbm}, CAT~\cite{cat2024}, and TabCBM~\cite{tabcbm2023}, mainly operate on homogeneous inputs or single-level concept structures. They do not provide dedicated encoding strategies for encrypted traffic, where evidence sources differ fundamentally in granularity, abstraction level, and temporal organization. Traffic-CBM therefore focuses on constructing a hierarchical concept interface for heterogeneous traffic evidence, including flow-level statistics, packet-level temporal behavior, packet-local byte patterns, and cross-packet byte regularities.

\subsection{Explainability in Encrypted Traffic Classification}

Compared with the rapid development of encrypted traffic classification itself, explainability research in this area remains limited and is still dominated by post-hoc interpretation. Existing studies typically use SHAP, LIME, attention visualization, or feature attribution to identify which flow features, packets, or input regions appear important for a given prediction. For example, Nascita \emph{et al.} use Deep SHAP to analyze multimodal mobile traffic classifiers~\cite{nascita2021xai}, while later work further applies post-hoc explanation in incremental learning settings~\cite{Nascita2023ExplainableTC,nascita2023xai}. Suh \emph{et al.} similarly adopt SHAP-based feature attribution for interpretable encrypted traffic detection without changing the predictive model itself~\cite{Suh2025SHAPTraffic}. Related patterns also appear in the broader network security literature~\cite{Benchama2023GANIDS,Hermosilla2025XAIIDS,Azam2023IDS}.

As a result, current explainability methods for encrypted traffic mainly reveal which raw features, packets, or input regions appear important, but rarely clarify how statistical, sequential, and byte-level evidence is organized once these heterogeneous modalities are fused inside a deep predictor~\cite{nascita2021xai,Nascita2023ExplainableTC,Suh2025SHAPTraffic}. In particular, few existing works explicitly model byte-level evidence as structured concept factors, let alone decompose it into packet-level and cross-packet factors. Unlike these SHAP/LIME-style post-hoc studies, Traffic-CBM builds interpretability into the model structure by organizing heterogeneous traffic evidence into an evidence-grounded hierarchical concept interface, enabling statistical, sequential, and byte-level signals to be analyzed within a unified concept-oriented framework~\cite{ji2025sinn}.
\section{Method}

\begin{figure*}[t]
    \centering
    \includegraphics[width=\textwidth]{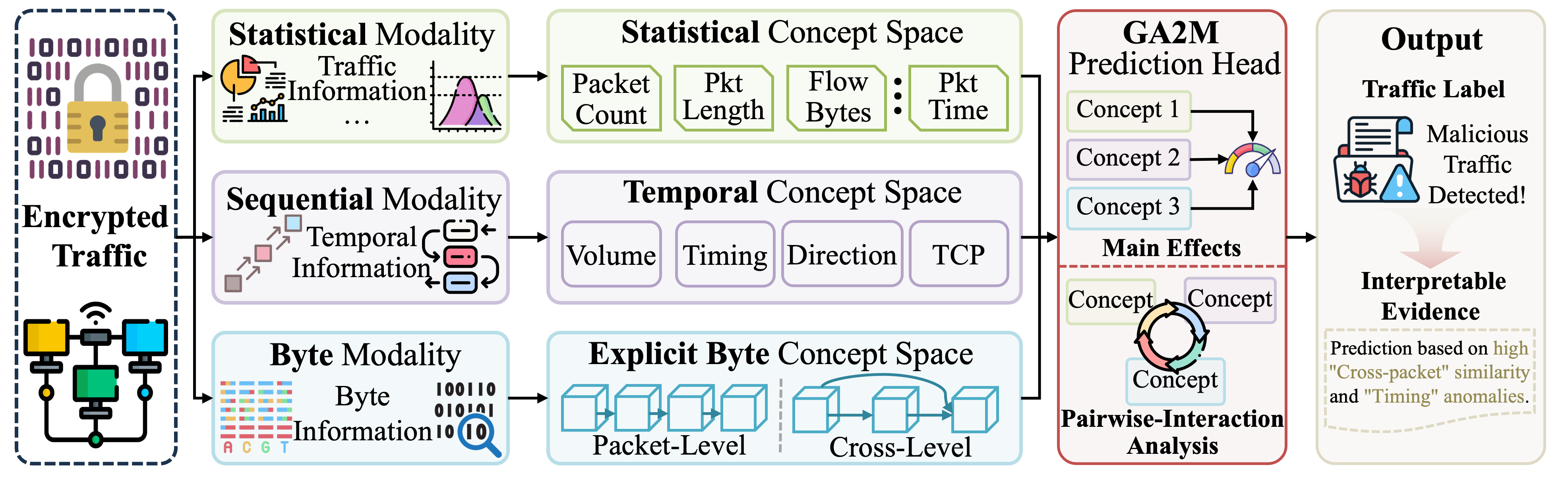}
    \vspace{-14pt}
    \caption{Overview of Traffic-CBM. Traffic-CBM organizes statistical, sequential, and byte-level traffic evidence into a hierarchical concept space composed of statistical concepts, temporal concepts, packet-level byte concepts, and cross-packet byte concepts. Final prediction is performed on this concept space through a GA2M head, enabling both main-effect and pairwise-interaction analysis.}
    \label{fig:overview}
    \vspace{-8pt}
\end{figure*}

\subsection{Overview}

We propose \textbf{Traffic-CBM}, a structurally interpretable multimodal framework for encrypted traffic classification. Instead of directly fusing heterogeneous traffic features into a single entangled latent representation, Traffic-CBM first organizes them into an explicitly structured concept space and then performs prediction on this concept space through an interaction-aware additive head.

For each traffic flow, we consider three input modalities:
\begin{equation}
x = \{x^{stat}, x^{seq}, x^{byte}\},
\end{equation}
where $x^{stat}$ denotes flow-level statistical features, $x^{seq}$ denotes packet-level sequential features, and $x^{byte}$ denotes tokenized packet bytes. These modalities are mapped into a unified concept representation:
\begin{equation}
c = [c^{stat}, c^{seq}, c^{pkt}, c^{cross}],
\end{equation}
where $c^{stat}$, $c^{seq}$, $c^{pkt}$, and $c^{cross}$ represent statistical, temporal, packet-level byte, and cross-packet byte concepts, respectively.

The concept space consists of 6 statistical concepts, 4 temporal concepts, 5 packet-level byte concepts, and 3 cross-packet byte concepts, resulting in an 18-dimensional representation. Unlike traditional concept bottleneck models that rely on explicit concept-value annotations~\cite{cbm2020}, Traffic-CBM defines concepts through traffic evidence constraints. Each concept is produced from a predefined and inspectable evidence source: statistical concepts are aligned with predefined feature groups, temporal concepts are learned from disjoint sequential subspaces, and byte-level concepts are separated into packet-level and cross-packet pathways. Traffic-CBM therefore follows the spirit of concept bottleneck modeling without requiring explicit concept supervision. Final prediction is performed through a generalized additive model with pairwise interactions (GA2M), yielding an interpretable prediction interface over the learned concept space.

\subsection{Input Representation}

Each traffic flow is represented by three modalities: flow-level statistical features, packet-level sequential features, and byte-level token sequences.

\noindent \textbf{Statistical modality.}
The statistical input $x^{stat}$ consists of normalized flow-level features extracted by CICFlowMeter~\cite{cicflowmeter}, describing macro traffic properties such as packet counts, length statistics, flow bytes, timing, TCP flags, and header/window information.

\noindent \textbf{Sequential modality.}
The sequential input $x^{seq} \in \mathbb{R}^{B \times N \times 7}$ is formed from the first $N$ packets of each flow in a batch of size $B$. Each packet is represented by seven features: packet length, inter-arrival time, direction, TCP flags, payload length, header length, and TCP window size. A binary attention mask is applied to distinguish valid packets from padded positions.

\noindent \textbf{Byte modality.}
The byte input $x^{byte} \in \mathbb{R}^{B \times 5 \times 63}$ is constructed from the first five packets of each flow. For each packet, a fixed-length byte window is extracted and converted into bigram tokens. This packet-wise organization preserves packet identity and positional locality, which explicitly supports subsequent packet-level and cross-packet byte concept modeling.

\subsection{Statistical Concept Modeling}

The statistical modality provides flow-level macro evidence for the hierarchical concept space. We partition the statistical features into six predefined groups corresponding to packet count, packet length, flow bytes, packet time, flag count, and header/window information. These groups are mapped to six scalar concepts:
\begin{equation}
c^{stat} = E_{stat}(x^{stat}).
\end{equation}

In practice, each statistical concept is produced by an independent multilayer perceptron operating on one predefined feature group. Let $x^{stat,(k)}$ denote the subvector corresponding to the $k$-th statistical group. Then the $k$-th concept is defined as
\begin{equation}
c^{stat}_{k} = \mathrm{MLP}_{k}(x^{stat,(k)}), \quad k=1,\dots,6.
\end{equation}

This grouped design improves evidence alignment, since each concept dimension is tied to an explicit flow-level factor rather than to a latent feature learned by a shared backbone.

\vspace{-5pt}
\subsection{Temporal Concept Modeling}

The sequential modality provides packet-level behavioral evidence. Instead of applying one shared encoder to all sequential features, Traffic-CBM uses dedicated temporal encoders over disjoint feature subspaces. Specifically, the seven packet-level features are divided into four groups corresponding to volume, timing, direction, and TCP-related behavior. These groups are mapped to four temporal concepts:
\begin{equation}
c^{seq} = E_{seq}(x^{seq}).
\end{equation}

For the $g$-th temporal group, we denote the corresponding input subset by $x^{seq,(g)}$ and write
\begin{equation}
c^{seq}_{g} = E^{(g)}_{seq}(x^{seq,(g)}), \quad g=1,\dots,4,
\end{equation}
where each $E^{(g)}_{seq}$ consists of an input projection, a group-specific Transformer encoder~\cite{vaswani2017transformer}, sequence pooling, and a scalar concept projection. Because the four temporal groups use separate projections, positional encodings, Transformer parameters, and concept heads, heterogeneous temporal evidence is modeled in separate encoders rather than being mixed within a single shared sequential encoder.

\subsection{Byte-level Concept Decomposition}

Instead of mapping all packet bytes into one entangled latent representation, Traffic-CBM explicitly decomposes byte-level evidence into packet-level and cross-packet concepts. The goal is to separate two distinct mechanisms in byte-level traffic evidence: localized intra-packet patterns and aligned inter-packet regularities.

\noindent \textbf{Packet-level byte concepts.}
We first define packet-level byte concepts, which summarize localized byte patterns within individual packets:
\begin{equation}
c^{pkt} = E_{pkt}(x^{byte}).
\end{equation}
In practice, $E_{pkt}$ is implemented as a shared packet-wise byte encoder applied to each packet separately. Let $x_i^{byte}$ denote the token sequence of the $i$-th packet. Then
\begin{equation}
c_i^{pkt} = E_{pkt}(x_i^{byte}), \quad i=1,\dots,5.
\end{equation}
Thus, the model preserves packet-specific byte evidence instead of collapsing all packet bytes into one global representation.

\noindent \textbf{Cross-packet byte concepts.}
We next define cross-packet byte concepts, which model aligned structural patterns across packets:
\begin{equation}
c^{cross} = E_{cross}(x^{byte}).
\end{equation}
Each packet is divided into three aligned local segments corresponding to the front, middle, and back parts of the token sequence. Corresponding segments across the five packets are then grouped together to form three cross-packet token sequences. Let $x^{cross,(g)}$ denote the token sequence of the $g$-th aligned cross-packet group. Then
\begin{equation}
c^{cross}_{g} = E_{cross}(x^{cross,(g)}), \quad g=1,\dots,3.
\end{equation}
Here, the same cross-packet encoder is shared across the three aligned groups. This design allows the model to capture whether structurally similar byte patterns recur at similar positions across packets.

Packet-level concepts capture localized intra-packet evidence, while cross-packet concepts capture shared regularities spanning multiple packets. A single shared byte encoder tends to entangle these two mechanisms, whereas explicit decomposition makes them separately analyzable within the concept space.

\subsection{Hierarchical Concept Representation}

The final concept representation is constructed by concatenating all concept groups:
\begingroup
\setlength{\abovedisplayskip}{4pt}
\setlength{\belowdisplayskip}{4pt} 
\begin{equation}
c = [c^{stat}, c^{seq}, c^{pkt}, c^{cross}].
\end{equation}
\endgroup
This yields a hierarchical concept space in which the four concept groups play complementary roles: $c^{stat}$ captures flow-level macro statistics, $c^{seq}$ captures temporal behavioral dynamics, $c^{pkt}$ captures localized packet-level byte cues, and $c^{cross}$ captures aligned cross-packet structural signals. Prediction is therefore performed on a compact and explicitly organized concept space rather than directly on entangled multimodal latent representations.

\subsection{Auxiliary Self-supervised Initialization}

To improve encoder initialization, we optionally pretrain the sequential and byte-level encoders before downstream supervised training. The sequential branch uses masked reconstruction on packet-level sequential features~\cite{mae2022}, while the byte branch uses masked token reconstruction on byte-level token sequences~\cite{bert2019}. Since self-supervised pretraining is not the focus of this work, we treat it as an auxiliary initialization strategy and study its effect separately in the ablation experiments.

\subsection{Prediction with GA2M}

Instead of predicting directly from raw heterogeneous features, Traffic-CBM performs classification through a generalized additive model with pairwise interactions (GA2M)~\cite{Lou2013GA2M,caruana2015ga2m} defined on the concept space. For a $K$-class classification task, the GA2M head produces a class-logit vector:
\begin{equation}
z = \beta_0 + \sum_i g_i(c_i) + \sum_{i<j} g_{ij}(c_i, c_j),
\end{equation}
where $z \in \mathbb{R}^{K}$ denotes the class-logit vector, $\beta_0 \in \mathbb{R}^{K}$ is the class-wise bias term, each $c_i$ is a scalar concept variable, $g_i(\cdot): \mathbb{R} \rightarrow \mathbb{R}^{K}$ models the class-wise individual contribution of concept $c_i$, and $g_{ij}(\cdot,\cdot): \mathbb{R}^{2} \rightarrow \mathbb{R}^{K}$ models the class-wise pairwise interaction between concepts $c_i$ and $c_j$. The final prediction is obtained by applying the softmax function to $z$.

In practice, $g_i$ and $g_{ij}$ are parameterized as shallow multilayer perceptrons, acting as neural shape functions over the concept space. This design makes the final decision process analyzable at two levels: the individual effect of each concept and the joint effect of concept pairs. By operating on scalar concept variables rather than raw high-dimensional features, the GA2M head provides an interpretable prediction interface over the structured concept space.

\section{Experiments}
\subsection{Experimental Setup}

\noindent \textbf{Datasets.} 
We evaluate Traffic-CBM on six encrypted traffic benchmarks: USTC-TFC2016, ISCX-VPN(App), ISCX-VPN(Service), CSTNET-TLS 1.3, CipherSpectrum(AES128), and CipherSpectrum(Mix). Table~\ref{tab:datasets} summarizes the datasets used in our experiments.

\begin{table}[t]
    \caption{Statistics of benchmark datasets.}
    \vspace{-8pt}
    \label{tab:datasets}
    \centering
    \begin{tabular*}{0.92\columnwidth}{@{\extracolsep{\fill}}ccc}
        \toprule
        Dataset & \#Flows & \#Classes \\
        \midrule
        CipherSpectrum(AES128)~\cite{sok2025}  & 41,000  & 41  \\
        CipherSpectrum(Mix)~\cite{sok2025}     & 41,205  & 41  \\
        CSTNET-TLS 1.3~\cite{etbert2022}         & 46,354  & 119 \\
        ISCX-VPN (App)~\cite{DraperGil2016}    & 2,259   & 15  \\
        ISCX-VPN (Service)~\cite{DraperGil2016}& 2,284   & 6   \\
        USTC-TFC2016~\cite{Wang2017MalwareCNN} & 190,229 & 16  \\
        \bottomrule
    \end{tabular*}
    
    \vspace{-10pt}
\end{table}

For all datasets, we use a unified 8:1:1 train/validation/test split. All methods are trained and evaluated on the same processed split, with consistent filtering and preprocessing criteria, discarding flows with fewer than three packets or those too short for reliable sequential or byte-level information. To reduce shortcut learning from label-revealing metadata, we apply leakage-controlled preprocessing before model input construction. Strong identification information (SII), including identifiers such as IP addresses, MAC addresses, and ports, is removed or sanitized. For CipherSpectrum, we additionally remove SNI-related information at the pcap/handshake level because SNI is plaintext TLS metadata and can directly expose the target hostname or service identifier~\cite{sok2025}. All baselines and Traffic-CBM are evaluated under the same leakage-controlled preprocessing protocol.

\noindent \textbf{Evaluation metrics.}
We report Accuracy (AC), Precision (PR), Recall (RC), and F1, and mainly use F1 as the primary metric since several datasets are class-imbalanced~\cite{zheng2020rbrn}.

\noindent \textbf{Baselines.}
We compare Traffic-CBM with representative encrypted traffic classification methods, including the traditional baseline AppScanner~\cite{appscanner2018}, earlier deep models DeepFP~\cite{deepfp2018} and FS-Net~\cite{fsnet2019}, and recent neural baselines ET-BERT~\cite{etbert2022}, YaTC~\cite{yatc2023}, TrafficFormer~\cite{trafficformer2025}, and NetMamba~\cite{netmamba2024}. These methods span several input paradigms, including flow statistics, packet-level sequences, and raw byte content.

\noindent \textbf{Implementation details.}
All experiments are conducted on a server with an NVIDIA RTX 4090 GPU. Unless otherwise specified, Traffic-CBM uses 6 statistical, 4 temporal, 5 packet-level byte, and 3 cross-packet concepts, together with a GA2M head and dropout 0.2. The sequential and byte encoders are 2-layer Transformers with model dimension 64 and 4 attention heads. We train for up to 80 epochs using AdamW~\cite{adamw2019} with batch size 256, learning rate $3\times10^{-4}$, weight decay $10^{-2}$, label smoothing 0.05, and early stopping patience 10. The sequential and byte-level encoders are initialized from self-supervised pre-training.

\subsection{Main Results}

We first evaluate whether Traffic-CBM remains competitive as an encrypted traffic classifier across diverse benchmarks.

\begin{table*}[t]
\centering
\caption{Results on CipherSpectrum(AES128), CipherSpectrum(Mix), and CSTNET-TLS 1.3 datasets.}
\vspace{-8pt}
\renewcommand{\arraystretch}{0.85}
\label{tab:results_part1}
\resizebox{\textwidth}{!}{%
\begin{tabular}{l|cccc|cccc|cccc}
\toprule

Datasets
& \multicolumn{4}{c|}{CipherSpectrum(AES128)}
& \multicolumn{4}{c|}{CipherSpectrum(Mix)}
& \multicolumn{4}{c}{CSTNET-TLS 1.3} \\

\midrule

Approaches
& AC & PR & RC & F1
& AC & PR & RC & F1
& AC & PR & RC & F1 \\

\midrule

AppScanner~\cite{appscanner2018}
& 0.9539 & 0.9566 & 0.9549 & 0.9553
& 0.9319 & 0.9332 & 0.9319 & 0.9320
& 0.5100 & 0.5456 & 0.4911 & 0.5023 \\

DeepFP~\cite{deepfp2018}
& 0.6098 & 0.6454 & 0.6098 & 0.6115
& 0.5995 & 0.6261 & 0.5995 & 0.5980
& 0.5114 & 0.5056 & 0.4827 & 0.4802 \\

FS-Net~\cite{fsnet2019}
& 0.9804 & 0.9810 & 0.9799 & 0.9802
& 0.9718 & 0.9728 & 0.9717 & 0.9720
& 0.8182 & 0.7914 & 0.7636 & 0.7652 \\

ET-BERT~\cite{etbert2022}
& 0.4512 & 0.4575 & 0.4512 & 0.3898
& 0.3280 & 0.3565 & 0.3280 & 0.2781
& 0.4610 & 0.4220 & 0.3992 & 0.3691 \\

YaTC~\cite{yatc2023}
& 0.4244 & 0.4199 & 0.4244 & 0.3648 
& 0.3702 & 0.3293 & 0.3702 & 0.3196
& 0.8923 & 0.8809 & 0.8761 & 0.8753 \\

TrafficFormer~\cite{trafficformer2025}
& 0.4159 & 0.3978 & 0.4159 & 0.3911
& 0.3249 & 0.3477 & 0.3249 & 0.2786
& 0.6592 & 0.6367 & 0.6194 & 0.6141 \\

NetMamba~\cite{netmamba2024}
& 0.4078 & 0.3582 & 0.4080 & 0.3435
& 0.3437 & 0.3029 & 0.3437 & 0.2859
& 0.8947 & 0.8980 & 0.8944 & 0.8926 \\

\midrule

Traffic-CBM
& 0.9602 & 0.9603 & 0.9602 & 0.9601
& 0.9329 & 0.9354 & 0.9329 & 0.9331
& 0.7581 & 0.7501 & 0.7242 & 0.7220 \\

\bottomrule
\end{tabular}
}
\vspace{-8pt}
\end{table*}

\begin{table*}[t]
\centering
\caption{Results on ISCX-VPN(App), ISCX-VPN(Service), and USTC-TFC2016 datasets.}
\vspace{-8pt}
\renewcommand{\arraystretch}{0.85}
\label{tab:results_part2}
\resizebox{\textwidth}{!}{%
\begin{tabular}{l|cccc|cccc|cccc}
\toprule

Datasets
& \multicolumn{4}{c|}{ISCX-VPN(App)}
& \multicolumn{4}{c|}{ISCX-VPN(Service)}
& \multicolumn{4}{c}{USTC-TFC2016} \\

\midrule

Approaches
& AC & PR & RC & F1
& AC & PR & RC & F1
& AC & PR & RC & F1 \\

\midrule

AppScanner~\cite{appscanner2018}
& 0.6041 & 0.5674 & 0.5789 & 0.5335
& 0.6522 & 0.7323 & 0.7957 & 0.7506
& 0.7433 & 0.8845 & 0.7177 & 0.7624 \\

DeepFP~\cite{deepfp2018}
& 0.3911 & 0.5927 & 0.3567 & 0.3689
& 0.7423 & 0.8169 & 0.7068 & 0.7474
& 0.6768 & 0.6453 & 0.6098 & 0.6115 \\

FS-Net~\cite{fsnet2019}
& 0.5000 & 0.3876 & 0.3715 & 0.3233
& 0.6836 & 0.4062 & 0.4618 & 0.4288
& 0.8446 & 0.9459 & 0.9322 & 0.9281 \\

ET-BERT~\cite{etbert2022}
& 0.6532 & 0.5523 & 0.5505 & 0.5341
& 0.7237 & 0.5787 & 0.6248 & 0.5985
& 0.9872 & 0.9889 & 0.9843 & 0.9865 \\

YaTC~\cite{yatc2023}
& 0.7867 & 0.7219 & 0.7102 & 0.7063
& 0.9258 & 0.9338 & 0.9310 & 0.9318
& 0.9916 & 0.9937 & 0.9886 & 0.9910 \\

TrafficFormer~\cite{trafficformer2025}
& 0.7162 & 0.6587 & 0.6693 & 0.6543
& 0.7763 & 0.7395 & 0.7601 & 0.7420
& 0.9872 & 0.9883 & 0.9852 & 0.9867 \\

NetMamba~\cite{netmamba2024}
& 0.8000 & 0.7891 & 0.8000 & 0.7877
& 0.8996 & 0.9032 & 0.8996 & 0.9004
& 0.9910 & 0.9910 & 0.9910 & 0.9909 \\

\midrule

Traffic-CBM
& 0.7928 & 0.7757 & 0.7097 & 0.7204
& 0.8991 & 0.8970 & 0.8800 & 0.8843
& 0.9914 & 0.9924 & 0.9898 & 0.9911 \\

\bottomrule
\end{tabular}
}
\vspace{-8pt}
\end{table*}

Tables~\ref{tab:results_part1} and~\ref{tab:results_part2} report the main comparison results on all benchmark datasets. Overall, Traffic-CBM achieves competitive performance across diverse encrypted traffic benchmarks while providing a clearer structural interpretation interface than conventional end-to-end fusion models.

Figure~\ref{fig:overall_balance} summarizes the overall trade-off across all six benchmarks. Traffic-CBM lies in the upper-left region of the plot, indicating a favorable trade-off between average Macro-F1 and cross-dataset variability.

Traffic-CBM is not the best-performing method on every dataset, but it remains competitive across all benchmarks and exhibits relatively consistent performance across datasets. This cross-dataset consistency is meaningful because the compared baselines rely on different types of traffic evidence, including raw byte content, flow statistics, and packet-level sequences. As the most informative signal can vary across datasets, methods built around a single evidence type may perform strongly on some benchmarks but less reliably on others.

In particular, Traffic-CBM performs strongly on the two CipherSpectrum benchmarks, reaching an F1 score of 0.9601 on CipherSpectrum(AES128) and 0.9331 on CipherSpectrum(Mix), suggesting that the proposed hierarchical concept space can effectively capture fine-grained byte-level and cross-packet evidence when such information is useful for discrimination.

Taken together, these results indicate that the proposed hierarchical concept space is expressive enough to support encrypted traffic classification across multiple datasets. Combined with the concept-space analyses presented later, Traffic-CBM is best understood as a structured and competitive model rather than a method designed solely to optimize raw predictive accuracy on every benchmark.

\subsection{Ablation Study of Key Components}

We next examine the contribution of the main architectural choices in Traffic-CBM. Table~\ref{tab:ablation_components} reports ablation results on representative datasets, focusing on the role of temporal concepts, packet-level byte concepts, cross-packet byte concepts, self-supervised pre-training, and the final prediction head.


\begin{table*}[t]
\centering
\caption{Ablation study of key components in Traffic-CBM on representative datasets.}
\vspace{-6pt}
\label{tab:ablation_components}
\setlength{\tabcolsep}{7pt}
\renewcommand{\arraystretch}{0.95}
\small
\begin{tabular}{cl ccccc *{4}{>{\centering\arraybackslash}p{1.4cm}}}
\toprule
\multicolumn{2}{c}{\multirow{2}{*}{\textbf{Method}}} 
& \multicolumn{5}{c}{\textbf{Components}} 
& \multicolumn{2}{c}{CipherSpectrum (AES128)} 
& \multicolumn{2}{c}{CSTNET-TLS 1.3} \\
\cmidrule(lr){3-7} \cmidrule(lr){8-9} \cmidrule(lr){10-11}
\multicolumn{2}{c}{} 
& Stat & Temp & Pkt & Cross & PT 
& AC & F1 & AC & F1 \\
\midrule

\multicolumn{2}{l}{\textbf{Traffic-CBM (full model)}} 
& $\checkmark$ & $\checkmark$ & $\checkmark$ & $\checkmark$ & $\checkmark$
& \textbf{0.9602} & \textbf{0.9601}
& \textbf{0.7581} & \textbf{0.7220} \\
\midrule

1 & Stat only
& $\checkmark$ & $\times$ & $\times$ & $\times$ & $\checkmark$
& 0.6661 & 0.6560
& 0.4966 & 0.4125 \\

2 & Stat + Temp
& $\checkmark$ & $\checkmark$ & $\times$ & $\times$ & $\checkmark$
& 0.8254 & 0.8220
& 0.6478 & 0.6054 \\

3 & Pkt + Cross
& $\times$ & $\times$ & $\checkmark$ & $\checkmark$ & $\checkmark$
& 0.4256 & 0.3912
& 0.5745 & 0.5185 \\

4 & Stat + Temp + Pkt
& $\checkmark$ & $\checkmark$ & $\checkmark$ & $\times$ & $\checkmark$
& 0.9322 & 0.9323
& 0.7503 & 0.7167 \\

5 & Stat + Temp + Cross
& $\checkmark$ & $\checkmark$ & $\times$ & $\checkmark$ & $\checkmark$
& 0.9466 & 0.9465
& 0.6842 & 0.6460 \\

\midrule

6 & w/ MLP head
& $\checkmark$ & $\checkmark$ & $\checkmark$ & $\checkmark$ & $\checkmark$
& 0.9632 & 0.9631
& 0.7973 & 0.7662 \\

7 & w/o Pre-training
& $\checkmark$ & $\checkmark$ & $\checkmark$ & $\checkmark$ & $\times$
& 0.9398 & 0.9399
& 0.7187 & 0.6821 \\
\bottomrule
\end{tabular}
\vspace{-10pt}
\end{table*}

First, adding temporal concepts consistently improves performance over the statistical-only model. For example, on CipherSpectrum(AES128), the F1 score increases from 0.6560 to 0.8220, and on CSTNET-TLS 1.3 it increases from 0.4125 to 0.6054. This indicates that packet-level temporal dynamics provide useful behavioral evidence beyond flow-level macro statistics.

Second, byte-level concepts further improve performance on top of the statistical+temporal backbone, but packet-level and cross-packet concepts contribute differently across datasets. On CipherSpectrum(AES128), \emph{Stat + Temp + Cross} outperforms \emph{Stat + Temp + Pkt} (0.9465 vs. 0.9323 F1), suggesting that cross-packet modeling is especially beneficial on this benchmark. By contrast, on CSTNET-TLS 1.3, \emph{Stat + Temp + Pkt} performs better than \emph{Stat + Temp + Cross}, indicating that packet-level byte evidence can be more useful on some datasets. The \emph{Pkt + Cross} byte-only variant performs substantially worse than multimodal variants, showing that byte-level evidence alone is much less effective without statistical and temporal context.

Third, self-supervised pre-training consistently improves the full model. Removing pre-training reduces F1 from 0.9601 to 0.9399 on CipherSpectrum(AES128) and from 0.7220 to 0.6821 on CSTNET-TLS 1.3, suggesting that self-supervised initialization benefits sequential and byte-level concept learning.

Finally, replacing the GA2M head with a two-layer MLP slightly improves performance on CipherSpectrum(AES128) (0.9631 vs. 0.9601 F1) and more substantially on CSTNET-TLS 1.3 (0.7662 vs. 0.7220 F1). This suggests a trade-off between predictive performance and structural analyzability: a more flexible black-box head can improve raw accuracy, while GA2M provides explicit main effects and pairwise concept interactions for concept-level analysis.

\vspace{-8pt}

\subsection{Concept-space Analysis}

Having established predictive performance and component utility, we next examine how the learned concept space contributes to the final decision process.

We first conduct detailed concept-space visualization on the CipherSpectrum(AES128) test set, where Traffic-CBM achieves strong performance and the fine-grained traffic categories make concept-level analysis especially informative. We adopt GA2M in the final model because it enables direct concept-level analysis of main effects and pairwise interactions. We later extend faithfulness and sufficiency analyses to additional datasets to examine whether the concept interface remains behaviorally meaningful beyond this favorable setting.

\noindent \textbf{Main Effects of Learned Concepts.}
To examine how the model uses the learned concept space, we first analyze the main effects of individual concepts in the GA2M head.

\begin{figure}[t]
    \centering
    \includegraphics[width=\columnwidth]{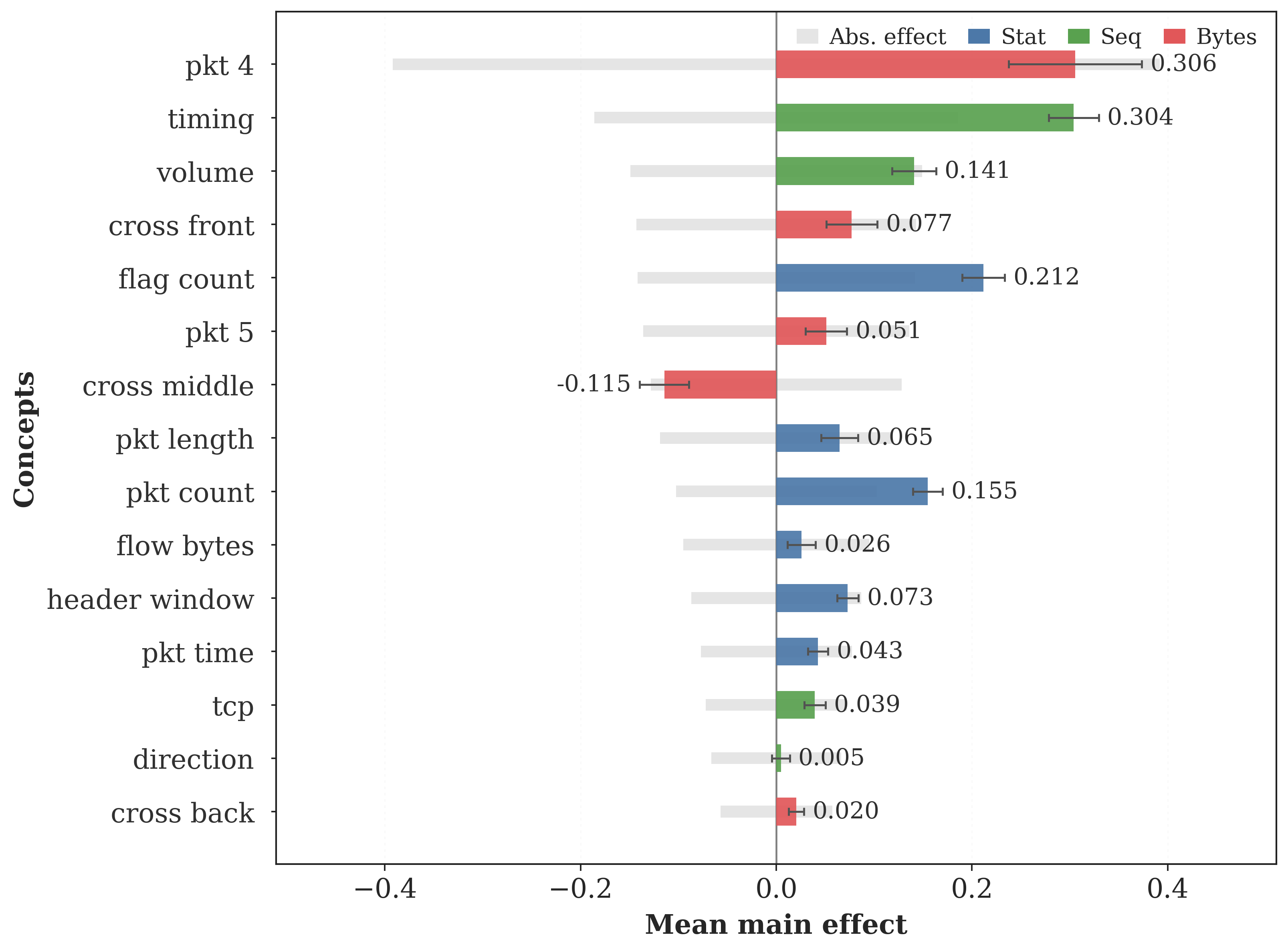}
    \vspace{-18pt}
    \caption{GA2M main effects of learned concepts on CipherSpectrum(AES128). Colored bars indicate signed mean main effects on the target logit, while the gray reference bands show the corresponding mean absolute effects. Concepts are sorted by mean absolute effect magnitude.}
    \label{fig:ga2m_main_effects}
    \vspace{-12pt}
\end{figure}

Figure~\ref{fig:ga2m_main_effects} shows the mean main effects of learned concepts on CipherSpectrum(AES128). The colored bars indicate the signed direction of the first-order effect, while the gray reference bands indicate the corresponding mean absolute effect magnitude. Several concepts from all three modalities appear among the strongest first-order terms, suggesting that the model does not rely on a single source of evidence.

In particular, packet-level concepts such as \emph{pkt 4}, together with sequential concepts such as \emph{timing} and \emph{volume}, statistical concepts such as \emph{flag count}, and byte-level concepts such as \emph{cross front}, exhibit strong main effects. This indicates that the learned concept space jointly captures flow-level macro properties, temporal behavioral dynamics, packet-level byte cues, and cross-packet byte patterns.

\noindent \textbf{Representative Shape Functions of Learned Concepts.}
To provide a finer-grained view of concept behavior, we further visualize the first-order shape functions of several representative concepts.

\begin{figure*}[t]
    \centering
    \includegraphics[width=\textwidth]{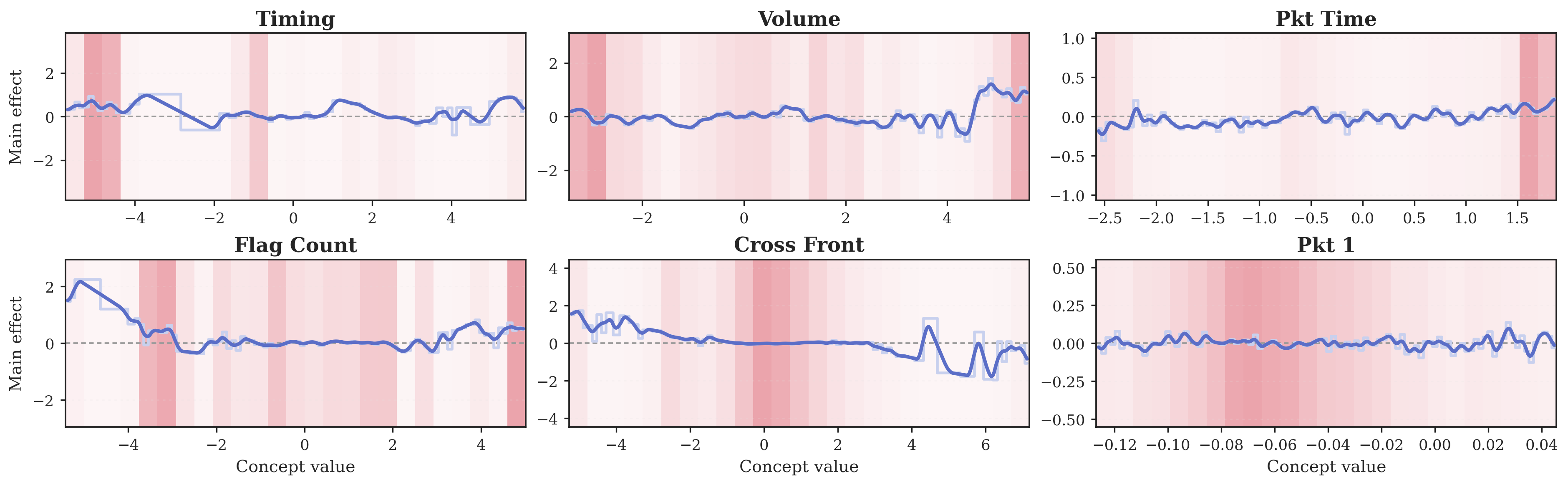}
    \vspace{-18pt}
    \caption{Representative first-order shape functions of selected learned concepts on CipherSpectrum(AES128). The x-axis denotes concept value, and the y-axis denotes the corresponding GA2M main effect. Background shading indicates the empirical distribution of concept values on the test set.}
    \label{fig:ga2m_shape_functions}
    \vspace{-12pt}
\end{figure*}

Figure~\ref{fig:ga2m_shape_functions} shows that different concepts influence the target logit in different ways across their value ranges. Statistical and sequential concepts such as \emph{timing}, \emph{volume}, \emph{pkt time}, and \emph{flag count} exhibit relatively smoother response patterns over their dominant value ranges, suggesting that part of the learned concept space corresponds to relatively stable first-order effects rather than highly entangled latent responses.

By contrast, byte-level concepts such as \emph{cross front} and \emph{pkt 1} show more localized and less smooth responses, which is consistent with their more structure-oriented semantics. These shape functions therefore complement the global ranking results by revealing not only which concepts are important, but also how they affect the prediction across different value ranges.

\noindent \textbf{Ranking of First- and Second-order Terms.}
Beyond first-order effects, a key advantage of GA2M is that it explicitly models second-order concept interactions. To compare the relative importance of main effects and pairwise interactions, we rank the strongest first-order and second-order GA2M terms by mean absolute effect.

\begin{figure}[t]
    \centering
    \includegraphics[width=\columnwidth]
    {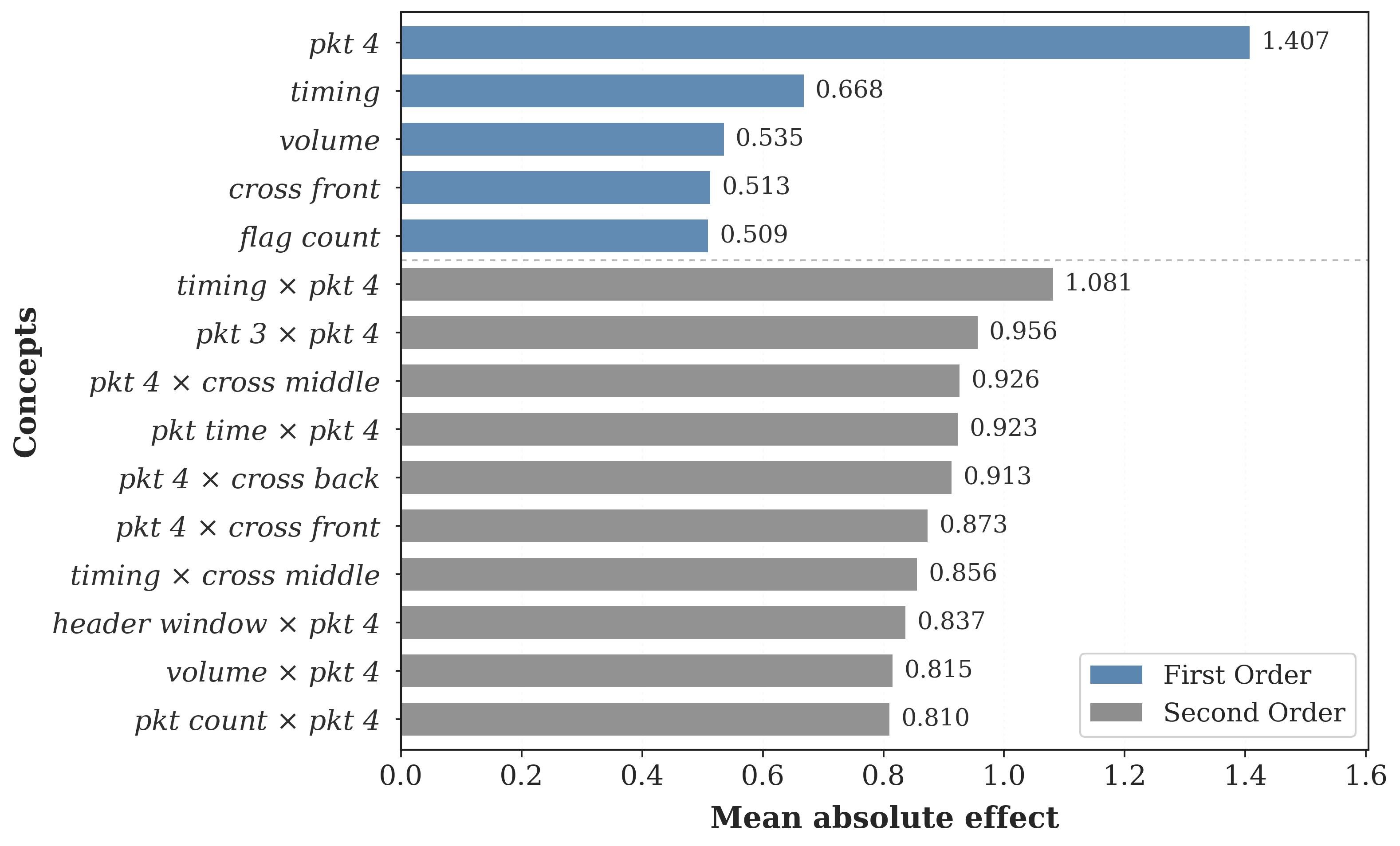}
    \vspace{-16pt}
    \caption{Ranking of the strongest first-order and second-order GA2M terms on CipherSpectrum(AES128), measured by mean absolute effect.}
    \label{fig:ga2m_term_ranking}
    \vspace{-14pt}
\end{figure}

Figure~\ref{fig:ga2m_term_ranking} shows that several second-order terms are stronger than many first-order terms, indicating that the final prediction is influenced not only by isolated concepts but also by their interactions. In particular, many of the strongest second-order terms are centered on the \emph{pkt 4} concept, such as \emph{timing $\times$ pkt 4}, \emph{pkt 3 $\times$ pkt 4}, \emph{pkt time $\times$ pkt 4}, \emph{volume $\times$ pkt 4}, and \emph{header window $\times$ pkt 4}. Strong interactions also appear between packet-level and cross-packet byte concepts, such as \emph{pkt 4 $\times$ cross middle}, \emph{pkt 4 $\times$ cross front}, and \emph{pkt 4 $\times$ cross back}, as well as between sequential and byte-level concepts such as \emph{timing $\times$ cross middle}.

Taken together, these results suggest that Traffic-CBM relies on a structured concept space in which first-order effects and cross-concept interactions jointly determine the final decision, rather than a simple additive list of independent concepts.

\subsection{Validation of Concept-level Attribution}

We next examine whether the concept-level attribution derived from the learned concept space is behaviorally meaningful, both within concept space and, for selected cases, at the input level.

\begin{table}[t]
\centering
\caption{Faithfulness evaluation via concept ablation. We progressively remove concepts based on their importance ranking and report the resulting performance.}
\vspace{-6pt}
\label{tab:faithfulness}
\setlength{\tabcolsep}{6pt}
\renewcommand{\arraystretch}{0.9}
\small
\begin{tabular}{lcccc}
\toprule
\textbf{Setting} & \textbf{\#Concepts} & \textbf{Accuracy} & \textbf{Macro-F1} & $\boldsymbol{\Delta}$\textbf{Acc} \\
\midrule
Full Model  & -- & 0.9602 & 0.9601 & -- \\
\midrule
Top-$k$     & 3  & 0.2483 & 0.2129 & $\downarrow$0.7119 \\
Bottom-$k$  & 3  & 0.9541 & 0.9540 & $\downarrow$0.0061 \\
\midrule
Top-$k$     & 5  & 0.1344 & 0.0977 & $\downarrow$0.8258 \\
Bottom-$k$  & 5  & 0.9554 & 0.9556 & $\downarrow$0.0048 \\
\midrule
Top-$k$     & 10 & 0.0295 & 0.0052 & $\downarrow$0.9307 \\
Bottom-$k$  & 10 & 0.8410 & 0.8330 & $\downarrow$0.1192 \\
\bottomrule
\end{tabular}
\vspace{-8pt}
\end{table}

\noindent \textbf{Faithfulness of Concept Attribution.}
To validate whether the learned concept importance reflects actual model behavior, we conduct a faithfulness analysis on the CipherSpectrum(AES128) test set by progressively ablating concepts based on their contribution ranking~\cite{deyoung2020eraser}.

As shown in Table~\ref{tab:faithfulness}, removing the top-$k$ most important concepts leads to a dramatic performance drop. For example, removing the top-5 concepts reduces the accuracy from 96.02\% to 13.44\%, indicating that the model prediction critically depends on a small subset of highly influential concepts. As $k$ increases, the degradation becomes even more severe, further suggesting that a small number of concepts dominate the decision process.

In contrast, removing the bottom-$k$ concepts has negligible impact on performance, suggesting that these concepts contribute little to the final prediction. The clear gap between top-$k$ and bottom-$k$ ablations indicates that the learned concept ranking is behaviorally meaningful rather than arbitrary.

Overall, these results suggest that Traffic-CBM relies on a structured and sparse concept space in which a small number of key concepts dominate prediction.

\begin{table}[t]
\centering
\caption{Sufficiency evaluation using only top-$k$ concepts on CipherSpectrum(AES128).}
\vspace{-7pt}
\label{tab:concept_sufficiency}
\setlength{\tabcolsep}{6pt}
\renewcommand{\arraystretch}{0.9}
\small
\begin{tabular}{lccc}
\toprule
\textbf{Setting} & \textbf{Accuracy} & \textbf{Macro-F1} & $\boldsymbol{\Delta}$\textbf{Acc} \\
\midrule
Full Model  & 0.9602 & 0.9601 & -- \\
\midrule
Top-3 Only  & 0.2924 & 0.2070 & $\downarrow$0.6678 \\
Top-5 Only  & 0.6012 & 0.5556 & $\downarrow$0.3590 \\
Top-10 Only & 0.9246 & 0.9239 & $\downarrow$0.0356 \\
\bottomrule
\end{tabular}
\vspace{-16pt}
\end{table}

\noindent \textbf{Sufficiency of Top-ranked Concepts.}
We further evaluate the sufficiency of the learned concept space by retaining only the top-$k$ concepts and removing the rest~\cite{deyoung2020eraser}.

As shown in Table~\ref{tab:concept_sufficiency}, using only the top-3 or top-5 concepts leads to a significant performance drop, while top-10 concepts already recover most of the performance. This indicates that although high-ranking concepts are critical, they are not individually sufficient.

Together with the faithfulness results, this suggests that the learned concept space encodes a structured decision process, where prediction emerges from the composition of multiple interdependent concepts rather than a small set of independent features.

\noindent \textbf{Cross-dataset Faithfulness.}
Tables~\ref{tab:faithfulness} and~\ref{tab:concept_sufficiency} provide detailed faithfulness and sufficiency analyses on CipherSpectrum(AES128). To examine whether the same trend holds across datasets, we further summarize the corresponding diagnostics on three representative datasets: CipherSpectrum(AES128), CSTNET-TLS13, and ISCX-VPN-Service.

\begin{table}[t]
\centering
\caption{Cross-dataset faithfulness and sufficiency evaluation. We compare the full model, ablation of the top-5 concepts, ablation of the bottom-5 concepts, and prediction using only the top-10 concepts.}
\vspace{-7pt}
\label{tab:cross_dataset_faithfulness}
\setlength{\tabcolsep}{4pt}
\renewcommand{\arraystretch}{0.95}
\small
\begin{tabular}{llcccc}
\toprule
\textbf{Dataset} & \textbf{Metric} & \textbf{Full} & \textbf{Top-5} & \textbf{Bottom-5} & \textbf{Top-10} \\
 & & & \textbf{Abl.} & \textbf{Abl.} & \textbf{Only} \\
\midrule
\multirow{2}{*}{Cipher-AES128}
& Acc. & 96.02 & 13.44 & 95.54 & 92.46 \\
& F1   & 96.01 & 9.77  & 95.56 & 92.39 \\
\midrule
\multirow{2}{*}{CSTNET-TLS13}
& Acc. & 75.81 & 19.80 & 69.37 & 57.54 \\
& F1   & 72.20 & 14.46 & 65.10 & 52.12 \\
\midrule
\multirow{2}{*}{ISCX-VPN-Service}
& Acc. & 89.91 & 41.67 & 87.28 & 82.46 \\
& F1   & 88.43 & 35.95 & 86.27 & 82.72 \\
\bottomrule
\end{tabular}
\vspace{-12pt}
\end{table}

As shown in Table~\ref{tab:cross_dataset_faithfulness}, ablating the top-ranked concepts causes substantially larger degradation than ablating the bottom-ranked concepts across all three datasets. This trend also holds on CSTNET-TLS13, a harder 119-class benchmark where Traffic-CBM is less competitive than the strongest black-box baselines: removing the top-5 concepts reduces accuracy from 75.81\% to 19.80\%, while removing the bottom-5 concepts only reduces it to 69.37\%.

The top-10-only results recover part, but not all, of the original performance. This suggests that high-ranking concepts retain substantial decision-relevant information, but the final prediction still relies on a structured combination of multiple concepts rather than a single dominant concept or shortcut cue.

\noindent \textbf{Input-level Validation of Timing Attribution.}
We next examine whether selected sample-wise attributions can be linked back to original traffic signals.

We conduct a timing-focused input perturbation study on the CipherSpectrum(AES128) test set by modifying packet inter-arrival times (IATs) in the original pcap while keeping byte content unchanged. We focus on the \emph{seq/timing} concept because it provides the most direct and controllable input-level validation pathway.

We construct two groups of high-confidence correctly classified samples according to the sample-wise contribution of \emph{seq/timing}: a \emph{timing-high} group, in which \emph{seq/timing} ranks within the top three concepts, and a \emph{timing-low} group, in which it falls within the bottom five. Each group contains three representative samples. For each sample, we apply random IAT perturbations at multiple jitter levels and average the results over three random seeds.

\begin{table}[t]
\centering
\caption{Group-level timing perturbation sensitivity on CipherSpectrum(AES128). Samples are divided according to the sample-wise importance of the \emph{seq/timing} concept, and results are averaged over three random seeds.}
\vspace{-7pt}
\label{tab:timing_attribution_validation}
\setlength{\tabcolsep}{6pt}
\renewcommand{\arraystretch}{0.9}
\small
\begin{tabular}{lcccc}
\toprule
\textbf{Group} & \textbf{Jitter} & \textbf{Avg. Conf.} & \textbf{$\Delta$Conf.} & \textbf{Flips} \\
\midrule
timing-high & 0.00 & 0.9773 & --                & 0/3 \\
timing-high & 0.03 & 0.9426 & $\downarrow$0.0366 & 0/9 \\
timing-high & 0.06 & 0.5671 & $\downarrow$0.4134 & 1/9 \\
timing-high & 0.09 & 0.6482 & $\downarrow$0.4481 & 4/9 \\
timing-high & 0.12 & 0.6949 & $\downarrow$0.4069 & 4/9 \\
\midrule
timing-low  & 0.00 & 0.9898 & --                & 0/3 \\
timing-low  & 0.03 & 0.9891 & $\downarrow$0.0012 & 0/9 \\
timing-low  & 0.06 & 0.9795 & $\downarrow$0.0203 & 0/9 \\
timing-low  & 0.09 & 0.9674 & $\downarrow$0.0382 & 0/9 \\
timing-low  & 0.12 & 0.9618 & $\downarrow$0.0326 & 0/9 \\
\bottomrule
\end{tabular}
\vspace{-10pt}
\end{table}

Table~\ref{tab:timing_attribution_validation} shows that the \emph{timing-high} samples are substantially more sensitive to timing perturbations than the \emph{timing-low} samples. For example, at jitter $=0.06$, the confidence drop is 0.4134 for the \emph{timing-high} group, compared with 0.0203 for the \emph{timing-low} group; at jitter $=0.09$, the corresponding drops are 0.4481 and 0.0382. Prediction flips also appear only in the \emph{timing-high} group under moderate perturbations.

These results provide strong input-level evidence that the learned \emph{seq/timing} attribution is grounded in real temporal cues and has a substantial impact on the model's decision-making process.

\noindent \textbf{Failure Inspection.}
We further inspect difficult failure cases to identify which evidence sources dominate incorrect predictions.

\begin{table}[t]
\centering
\caption{Representative failure cases on CSTNET-TLS13.}
\vspace{-7pt}
\label{tab:failure_cases}
\setlength{\tabcolsep}{3.5pt}
\renewcommand{\arraystretch}{0.95}
\small
\begin{tabular}{llcl}
\toprule
\textbf{True} & \textbf{Pred.} & \textbf{Conf.} & \textbf{Top Concepts} \\
\midrule
\texttt{gravatar} & \texttt{twitter} & 0.4426 &
cross-mid, pkt-1, pkt-3, dir., cross-back \\
\texttt{netflix} & \texttt{yahoo} & 0.6078 &
pkt-len, pkt-2, pkt-3, flag-cnt, cross-front \\
\bottomrule
\end{tabular}
\vspace{-10pt}
\end{table}

Table~\ref{tab:failure_cases} shows representative misclassified samples on CSTNET-TLS13. Here, \emph{cross-mid}, \emph{dir.}, \emph{pkt-len}, and \emph{flag-cnt} denote cross-middle, direction, packet-length, and flag-count concepts, respectively. For example, when a \emph{gravatar.com} sample is misclassified as \emph{twitter.com}, the dominant contributions come from cross-packet byte concepts, packet-level byte concepts, and direction-related evidence. When a \emph{netflix.com} sample is misclassified as \emph{yahoo.com}, the decision is mainly driven by packet-length, packet-level byte concepts, flag-count, and cross-packet byte evidence.

These cases further show that the concept interface can be used to inspect not only correct predictions but also difficult failure modes by exposing which evidence pathways dominate the model's decisions.

\section{Conclusion}

In this paper, we presented Traffic-CBM, a structurally interpretable multimodal framework for encrypted traffic classification. Instead of directly fusing heterogeneous traffic evidence into a black-box representation, Traffic-CBM organizes statistical, sequential, and byte-level evidence into a unified hierarchical concept space and performs prediction through a GA2M-based head. Experiments on multiple encrypted traffic benchmarks show that Traffic-CBM achieves competitive and balanced classification performance while supporting concept-level analysis of the prediction process. Further analyses, including faithfulness and sufficiency evaluation, input-level timing perturbation, and failure inspection, show that the learned concept space provides an inspectable interface for understanding how different traffic evidence sources contribute to model decisions. Overall, our results suggest that encrypted traffic classification can benefit from moving beyond purely post-hoc explanation toward explicitly structured concept-based modeling. In future work, we plan to explore broader concept validation, more expressive yet still analyzable prediction heads, and human-centered evaluation of concept-based explanations for traffic analysis.

\clearpage
\bibliographystyle{ACM-Reference-Format}
\bibliography{main}

@inproceedings{zheng2020rbrn,
  author = {Zheng, Wenbo and Gou, Chao and Yan, Lan and Mo, Shaocong},
  title = {Learning to Classify: A Flow-Based Relation Network for Encrypted Traffic Classification},
  booktitle = {Proceedings of the ACM Web Conference (WWW)},
  pages = {13--22},
  year = {2020},
  doi = {10.1145/3366423.3380090}
}

@inproceedings{flowprint2020,
  author = {van Ede, Thijs Sebastiaan and Riccardo Bortolameotti and Andrea Continella and Jingjing Ren and Dubois, Daniel J. and Martina Lindorfer and David Choffnes and van Steen, Maarten and Andreas Peter},
  title = {FlowPrint: Semi-Supervised Mobile-App Fingerprinting on Encrypted Network Traffic},
  booktitle = {Network and Distributed System Security Symposium (NDSS)},
  year = {2020},
  doi = {10.14722/ndss.2020.24412}
}

@article{appscanner2018,
  author = {Taylor, Vincent F. and Spolaor, Riccardo and Conti, Mauro and Martinovic, Ivan},
  title = {Robust Smartphone App Identification via Encrypted Network Traffic Analysis},
  journal = {IEEE Transactions on Information Forensics and Security},
  volume = {13},
  number = {1},
  pages = {63--78},
  year = {2018},
  doi = {10.1109/TIFS.2017.2737970}
}

@inproceedings{hayes2016kfp,
  author = {Hayes, Jamie and Danezis, George},
  title = {k-fingerprinting: a robust scalable website fingerprinting technique},
  booktitle = {Proceedings of the USENIX Security Symposium},
  pages = {1187--1203},
  year = {2016}
}

@inproceedings{alnaami2016bind,
  author = {Al-Naami, Khaled and Chandra, Swarup and Mustafa, Ahmad and Khan, Latifur and Lin, Zhiqiang and Hamlen, Kevin and Thuraisingham, Bhavani},
  title = {Adaptive Encrypted Traffic Fingerprinting with Bi-Directional Dependence},
  booktitle = {Proceedings of the Annual Computer Security Applications Conference (ACSAC)},
  pages = {177--188},
  year = {2016}
}

@inproceedings{deepfp2018,
  author = {Sirinam, Payap and Imani, Mohsen and Juarez, Marc and Wright, Matthew},
  title = {Deep Fingerprinting: Undermining Website Fingerprinting Defenses with Deep Learning},
  booktitle = {Proceedings of the ACM SIGSAC Conference on Computer and Communications Security (CCS)},
  pages = {1928--1943},
  year = {2018},
  doi = {10.1145/3243734.3243768}
}

@inproceedings{fsnet2019,
  author = {Liu, Chang and He, Longtao and Xiong, Gang and Cao, Zigang and Li, Zhen},
  title = {FS-Net: A Flow Sequence Network for Encrypted Traffic Classification},
  booktitle = {Proceedings of the IEEE Conference on Computer Communications (INFOCOM)},
  pages = {1171--1179},
  year = {2019},
  doi = {10.1109/INFOCOM.2019.8737507}
}

@article{deeppacket2020,
  author = {Lotfollahi, Mohammad and Jafari Siavoshani, Mahdi and Shirali Hossein Zade, Ramin and Saberian, Mohammdsadegh},
  title = {Deep Packet: A Novel Approach for Encrypted Traffic Classification Using Deep Learning},
  journal = {Soft Computing},
  volume = {24},
  number = {3},
  pages = {1999--2012},
  year = {2020},
  doi = {10.1007/s00500-019-04030-2}
}

@article{tscrnn2021,
  author = {Kunda Lin and Xiaolong Xu and Honghao Gao},
  title = {TSCRNN: A Novel Classification Scheme of Encrypted Traffic Based on Flow Spatiotemporal Features for Efficient Management of IIoT},
  journal = {Computer Networks},
  volume = {190},
  pages = {107974},
  year = {2021},
  doi = {10.1016/j.comnet.2021.107974}
}

@article{swift2025,
  author = {Xi, Tieqi and Zheng, Qiuhua and Cheng, Chuanhui and Wu, Ting and Xie, Guojie and Qian, Xuebiao and Ye, Haochen and Sun, Zhenyu},
  title = {SwiftSession: A Novel Incremental and Adaptive Approach to Rapid Traffic Classification by Leveraging Local Features},
  journal = {Future Internet},
  volume = {17},
  number = {3},
  pages = {114},
  year = {2025},
  doi = {10.3390/fi17030114}
}

@article{graphdapp2021,
  author = {Shen, Meng and Zhang, Jinpeng and Zhu, Liehuang and Xu, Ke and Du, Xiaojiang},
  title = {Accurate Decentralized Application Identification via Encrypted Traffic Analysis Using Graph Neural Networks},
  journal = {IEEE Transactions on Information Forensics and Security},
  volume = {16},
  pages = {2367--2380},
  year = {2021},
  doi = {10.1109/TIFS.2021.3050608}
}

@inproceedings{pert2020,
  author = {He, Hong Ye and Guo Yang, Zhi and Chen, Xiang Ning},
  title = {PERT: Payload Encoding Representation from Transformer for Encrypted Traffic Classification},
  booktitle = {2020 ITU Kaleidoscope: Industry-Driven Digital Transformation (ITU K)},
  pages = {1--8},
  year = {2020},
  doi = {10.23919/ITUK50268.2020.9303204}
}

@inproceedings{etbert2022,
  author = {Lin, Xinjie and Xiong, Gang and Gou, Gaopeng and Li, Zhen and Shi, Junzheng and Yu, Jing},
  title = {ET-BERT: A Contextualized Datagram Representation with Pre-training Transformers for Encrypted Traffic Classification},
  booktitle = {Proceedings of the ACM Web Conference (WWW)},
  pages = {633--642},
  year = {2022},
  doi = {10.1145/3485447.3512217}
}

@inproceedings{trafficformer2025,
  author = {Zhou, Guangmeng and Guo, Xiongwen and Liu, Zhuotao and Li, Tong and Li, Qi and Xu, Ke},
  title = {TrafficFormer: An Efficient Pre-trained Model for Traffic Data},
  booktitle = {Proceedings of the IEEE Symposium on Security and Privacy (SP)},
  pages = {1844--1860},
  year = {2025},
  doi = {10.1109/SP61157.2025.00102}
}

@article{yatc2023,
  author = {Zhao, Ruijie and Zhan, Mingwei and Deng, Xianwen and Wang, Yanhao and Wang, Yijun and Gui, Guan and Xue, Zhi},
  title = {Yet Another Traffic Classifier: A Masked Autoencoder Based Traffic Transformer with Multi-Level Flow Representation},
  journal = {Proceedings of the AAAI Conference on Artificial Intelligence},
  volume = {37},
  number = {4},
  pages = {5420--5427},
  year = {2023},
  doi = {10.1609/aaai.v37i4.25674}
}

@inproceedings{netmamba2024,
  author = {Wang, Tongze and Xie, Xiaohui and Wang, Wenduo and Wang, Chuyi and Zhao, Youjian and Cui, Yong},
  title = {NetMamba: Efficient Network Traffic Classification via Pre-Training Unidirectional Mamba},
  booktitle = {Proceedings of the IEEE International Conference on Network Protocols (ICNP)},
  pages = {1--11},
  year = {2024},
  doi = {10.1109/ICNP61940.2024.10858569}
}

@inproceedings{mm4flow2025,
  author = {Yang, Luming and Liu, Lin and Huang, JunJie and Liu, Zhuotao and Liang, Shiyu and Fu, Shaojing and Wang, Yongjun},
  title = {MM4flow: A Pre-trained Multi-modal Model for Versatile Network Traffic Analysis},
  booktitle = {Proceedings of the ACM SIGSAC Conference on Computer and Communications Security (CCS)},
  pages = {1664--1678},
  year = {2025},
  doi = {10.1145/3719027.3744804}
}

@inproceedings{flowmae2023,
  author = {Hang, Zijun and Lu, Yuliang and Wang, Yongjie and Xie, Yi},
  title = {Flow-MAE: Leveraging Masked AutoEncoder for Accurate, Efficient and Robust Malicious Traffic Classification},
  booktitle = {Proceedings of the International Symposium on Research in Attacks, Intrusions and Defenses (RAID)},
  pages = {297--314},
  year = {2023},
  doi = {10.1145/3607199.3607206}
}

@inproceedings{jia2025pira,
  author = {Jia, Haozhe and Chen, Wenshuo and Huang, Zhihui and Wang, Lei and Xiao, Hongru and Jia, Nanqian and Wu, Keming and Lai, Songning and Tian, Bowen and Yue, Yutao},
  title = {Physics-Informed Representation Alignment for Sparse Radio-Map Reconstruction},
  booktitle = {Proceedings of the ACM International Conference on Multimedia (ACM MM)},
  pages = {12352--12360},
  year = {2025},
  doi = {10.1145/3746027.3758161}
}

@inproceedings{tian2025text2weight,
  author = {Tian, Bowen and Chen, Wenshuo and Li, Zexi and Lai, Songning and Wu, Jiemin and Yue, Yutao},
  title = {Text2Weight: Bridging Natural Language and Neural Network Weight Spaces},
  booktitle = {Proceedings of the ACM International Conference on Multimedia (ACM MM)},
  pages = {10152--10160},
  year = {2025},
  doi = {10.1145/3746027.3755441}
}

@inproceedings{cbm2020,
  author = {Koh, Pang Wei and Nguyen, Thao and Tang, Yew Siang and Mussmann, Stephen and Pierson, Emma and Kim, Been and Liang, Percy},
  title = {Concept Bottleneck Models},
  booktitle = {Proceedings of the International Conference on Machine Learning (ICML)},
  pages = {5338--5348},
  year = {2020}
}

@inproceedings{oikarinen2023cbm,
  author = {Tuomas Oikarinen and Subhro Das and Lam M. Nguyen and Tsui-Wei Weng},
  title = {Label-free Concept Bottleneck Models},
  booktitle = {The Eleventh International Conference on Learning Representations},
  year = {2023}
}

@inproceedings{cat2024,
  author = {Duong, Viet and Wu, Qiong and Zhou, Zhengyi and Zhao, Hongjue and Luo, Chenxiang and Zavesky, Eric and Yao, Huaxiu and Shao, Huajie},
  title = {CAT: Interpretable Concept-based Taylor Additive Models},
  booktitle = {Proceedings of the 30th ACM SIGKDD Conference on Knowledge Discovery and Data Mining (KDD)},
  pages = {723--734},
  year = {2024},
  doi = {10.1145/3637528.3672020}
}

@inproceedings{Lou2013GA2M,
  author = {Lou, Yin and Caruana, Rich and Gehrke, Johannes and Hooker, Giles},
  title = {Accurate Intelligible Models with Pairwise Interactions},
  booktitle = {Proceedings of the 19th ACM SIGKDD International Conference on Knowledge Discovery and Data Mining (KDD)},
  year = {2013},
  pages = {623--631},
  doi = {10.1145/2487575.2487579}
}

@article{tabcbm2023,
  author = {Mateo Espinosa Zarlenga and Zohreh Shams and Michael Edward Nelson and Been Kim and Mateja Jamnik},
  title = {TabCBM: Concept-based Interpretable Neural Networks for Tabular Data},
  journal = {Transactions on Machine Learning Research},
  year = {2023}
}

@inproceedings{hybridcbm2025,
  author = {Liu, Yang and Zhang, Tianwei and Gu, Shi},
  title = {Hybrid Concept Bottleneck Models},
  booktitle = {Proceedings of the IEEE/CVF Conference on Computer Vision and Pattern Recognition (CVPR)},
  pages = {20179--20189},
  year = {2025},
  doi = {10.1109/CVPR52734.2025.01879}
}

@article{poeta2025cxai,
  author = {Poeta, Eleonora and Ciravegna, Gabriele and Pastor, Eliana and Cerquitelli, Tania and Baralis, Elena},
  title = {Concept-based Explainable Artificial Intelligence: A Survey},
  journal = {ACM Computing Surveys},
  year = {2025},
  doi = {10.1145/3774643}
}

@inproceedings{agarwal2021nam,
  author = {Rishabh Agarwal and Levi Melnick and Nicholas Frosst and Xuezhou Zhang and Ben Lengerich and Rich Caruana and Geoffrey Hinton},
  title = {Neural Additive Models: Interpretable Machine Learning with Neural Nets},
  booktitle = {Advances in Neural Information Processing Systems (NeurIPS)},
  year = {2021}
}

@article{ji2025sinn,
  author = {Ji, Yang and Sun, Ying and Zhang, Yuting and Wang, Zhigaoyuan and Zhuang, Yuanxin and Gong, Zheng and Shen, Dazhong and Qin, Chuan and Zhu, Hengshu and Xiong, Hui},
  title = {A Comprehensive Survey on Self-Interpretable Neural Networks},
  journal = {Proceedings of the IEEE},
  volume = {113},
  pages = {783--813},
  year = {2025},
  doi = {10.1109/JPROC.2025.3635153}
}

@inproceedings{caruana2015ga2m,
  author = {Caruana, Rich and Lou, Yin and Gehrke, Johannes and Koch, Paul and Sturm, Marc and Elhadad, Noemie},
  title = {Intelligible Models for HealthCare: Predicting Pneumonia Risk and Hospital 30-day Readmission},
  booktitle = {Proceedings of the ACM SIGKDD International Conference on Knowledge Discovery and Data Mining (KDD)},
  pages = {1721--1730},
  year = {2015},
  doi = {10.1145/2783258.2788613}
}

@inproceedings{deyoung2020eraser,
  author = {DeYoung, Jay and Jain, Sarthak and Rajani, Nazneen Fatema and Lehman, Eric and Xiong, Caiming and Socher, Richard and Wallace, Byron C.},
  title = {ERASER: A Benchmark to Evaluate Rationalized NLP Models},
  booktitle = {Proceedings of the Annual Meeting of the Association for Computational Linguistics (ACL)},
  pages = {4443--4458},
  year = {2020},
  doi = {10.18653/v1/2020.acl-main.408}
}

@article{Azam2023IDS,
  author = {Azam, Zahedi and Islam, Md. Motaharul and Huda, Mohammad Nurul},
  title = {Comparative Analysis of Intrusion Detection Systems and Machine Learning-Based Model Analysis Through Decision Tree},
  journal = {IEEE Access},
  volume = {11},
  pages = {80348--80391},
  year = {2023},
  doi = {10.1109/ACCESS.2023.3296444}
}

@article{Benchama2023GANIDS,
  author = {Benchama, Asmaa and Zebbara, Khalid},
  title = {Novel Approach to Intrusion Detection: Introducing GAN-MSCNN-BILSTM with LIME Predictions},
  journal = {Data and Metadata},
  volume = {2},
  year = {2023},
  doi = {10.56294/dm2023202}
}

@article{Hermosilla2025XAIIDS,
  author = {Hermosilla, Pamela and Berríos, Sebastián and Allende-Cid, Héctor},
  title = {Explainable AI for Forensic Analysis: A Comparative Study of SHAP and LIME in Intrusion Detection Models},
  journal = {Applied Sciences},
  volume = {15},
  number = {13},
  year = {2025},
  doi = {10.3390/app15137329}
}

@inproceedings{Suh2025SHAPTraffic,
  author = {Suh, Jiwon and Hong, Juwon and Gu, Mose and Jeong, Jaehoon Paul},
  title = {Interpretable Detection of Encrypted Traffic Using SHAP-Based Feature Attribution},
  booktitle = {Proceedings of the International Conference on Information and Communication Technology Convergence (ICTC)},
  year = {2025},
  pages = {190--195},
  doi = {10.1109/ICTC66702.2025.11388509}
}

@inproceedings{Nascita2023ExplainableTC,
  author = {Nascita, Alfredo and Cerasuolo, Francesco and Aceto, Giuseppe and Ciuonzo, Domenico and Persico, Valerio and Pescap\'{e}, Antonio},
  title = {Explainable Mobile Traffic Classification: The Case of Incremental Learning},
  booktitle = {Proceedings of the 2023 Workshop on Explainable and Safety Bounded Machine Learning for Networking},
  year = {2023},
  pages = {25--31},
  doi = {10.1145/3630050.3630178}
}

@article{nascita2021xai,
  author = {Nascita, Alfredo and Montieri, Antonio and Aceto, Giuseppe and Ciuonzo, Domenico and Persico, Valerio and Pescapé, Antonio},
  title = {XAI Meets Mobile Traffic Classification: Understanding and Improving Multimodal Deep Learning Architectures},
  journal = {IEEE Transactions on Network and Service Management},
  year = {2021},
  volume = {18},
  number = {4},
  pages = {4225--4246},
  doi = {10.1109/TNSM.2021.3098157}
}

@article{nascita2023xai,
  author = {Nascita, Alfredo and Montieri, Antonio and Aceto, Giuseppe and Ciuonzo, Domenico and Persico, Valerio and Pescap\'{e}, Antonio},
  title = {Improving Performance, Reliability, and Feasibility in Multimodal Multitask Traffic Classification with XAI},
  journal = {IEEE Transactions on Network and Service Management},
  volume = {20},
  pages = {1267--1289},
  year = {2023},
  doi = {10.1109/TNSM.2023.3246794}
}

@inproceedings{DraperGil2016,
  author = {Gerard Draper-Gil and Arash Habibi Lashkari and Mohammad Saiful Islam Mamun and Ali A. Ghorbani},
  title = {Characterization of Encrypted and VPN Traffic Using Time-Related Features},
  booktitle = {Proceedings of the International Conference on Information Systems Security and Privacy (ICISSP)},
  year = {2016}
}

@inproceedings{Wang2017MalwareCNN,
  author = {Wei Wang and Ming Zhu and Xuewen Zeng and Xiaozhou Ye and Yiqiang Sheng},
  title = {Malware Traffic Classification Using Convolutional Neural Network for Representation Learning},
  booktitle = {Proceedings of the International Conference on Information Networking (ICOIN)},
  year = {2017},
  pages = {712--717},
  doi = {10.1109/ICOIN.2017.7899588}
}

@inproceedings{sok2025,
  author = {Wickramasinghe, Nimesha and Shaghaghi, Arash and Tsudik, Gene and Jha, Sanjay},
  title = {SoK: Decoding the Enigma of Encrypted Network Traffic Classifiers},
  booktitle = {Proceedings of the IEEE Symposium on Security and Privacy (SP)},
  pages = {1825--1843},
  year = {2025},
  doi = {10.1109/SP61157.2025.00165}
}

@article{Dong2025Review,
  author = {Wenqi Dong and Jing Yu and Xinjie Lin and Gaopeng Gou and Gang Xiong},
  title = {Deep learning and pre-training technology for encrypted traffic classification: A comprehensive review},
  journal = {Neurocomputing},
  volume = {617},
  pages = {128444},
  year = {2025},
  doi = {10.1016/j.neucom.2024.128444}
}

@article{Sharma2025Survey,
  author = {Sharma, Adit and Lashkari, Arash Habibi},
  title = {A Survey on Encrypted Network Traffic: Identification and Classification Techniques, Challenges, and Future Directions},
  journal = {Computer Networks},
  volume = {257},
  pages = {110984},
  year = {2025},
  doi = {10.1016/j.comnet.2024.110984}
}

@inproceedings{cicflowmeter,
  author = {Iman Sharafaldin and Arash Habibi Lashkari and Ali A. Ghorbani},
  title = {Toward Generating a New Intrusion Detection Dataset and Intrusion Traffic Characterization},
  booktitle = {Proceedings of the International Conference on Information Systems Security and Privacy (ICISSP)},
  year = {2018}
}

@inproceedings{bert2019,
  author = {Jacob Devlin and Ming-Wei Chang and Kenton Lee and Kristina Toutanova},
  title = {BERT: Pre-training of Deep Bidirectional Transformers for Language Understanding},
  booktitle = {Proceedings of the North American Chapter of the Association for Computational Linguistics (NAACL)},
  year = {2019}
}

@inproceedings{lime2016,
  author = {Ribeiro, Marco Tulio and Singh, Sameer and Guestrin, Carlos},
  title = {"Why Should I Trust You?": Explaining the Predictions of Any Classifier},
  booktitle = {Proceedings of the ACM SIGKDD International Conference on Knowledge Discovery and Data Mining (KDD)},
  pages = {1135--1144},
  year = {2016},
  doi = {10.1145/2939672.2939778}
}

@inproceedings{shap2017,
  author = {Lundberg, Scott M. and Lee, Su-In},
  title = {A Unified Approach to Interpreting Model Predictions},
  booktitle = {Advances in Neural Information Processing Systems (NeurIPS)},
  pages = {4768--4777},
  year = {2017}
}

@inproceedings{adamw2019,
  author = {Ilya Loshchilov and Frank Hutter},
  title = {Decoupled Weight Decay Regularization},
  booktitle = {International Conference on Learning Representations (ICLR)},
  year = {2019}
}

@inproceedings{mae2022,
  author = {He, Kaiming and Chen, Xinlei and Xie, Saining and Li, Yanghao and Dollár, Piotr and Girshick, Ross},
  title = {Masked Autoencoders Are Scalable Vision Learners},
  booktitle = {Proceedings of the IEEE/CVF Conference on Computer Vision and Pattern Recognition (CVPR)},
  pages = {15979--15988},
  year = {2022},
  doi = {10.1109/CVPR52688.2022.01553}
}

@inproceedings{vaswani2017transformer,
  author = {Vaswani, Ashish and Shazeer, Noam and Parmar, Niki and Uszkoreit, Jakob and Jones, Llion and Gomez, Aidan N. and Kaiser, \L{}ukasz and Polosukhin, Illia},
  title = {Attention Is All You Need},
  booktitle = {Advances in Neural Information Processing Systems (NeurIPS)},
  pages = {6000--6010},
  year = {2017}
}

\clearpage
\appendix
\section{Dataset Details}

We use six public encrypted traffic benchmarks in this work: CipherSpectrum(AES128), CipherSpectrum(Mix), CSTNET-TLS 1.3, ISCX-VPN(App), ISCX-VPN(Service), and USTC-TFC2016~\cite{sok2025,etbert2022,DraperGil2016,Wang2017MalwareCNN}. Together, these datasets cover several representative encrypted traffic classification settings, including application-level classification, service-level classification, TLS 1.3 web traffic classification, and malicious traffic classification.

\noindent \textbf{CipherSpectrum(AES128) and CipherSpectrum(Mix).}
CipherSpectrum is a contemporary encrypted traffic dataset designed to reflect TLS 1.3 traffic conditions more faithfully~\cite{sok2025}. In our experiments, we use two benchmark variants derived from this dataset, namely CipherSpectrum(AES128) and CipherSpectrum(Mix). These two variants serve as the main TLS 1.3 benchmarks in our study and are used to evaluate whether the proposed model can capture fine-grained byte-level and cross-packet evidence.

\noindent \textbf{CSTNET-TLS 1.3.}
CSTNET-TLS 1.3 is a benchmark for encrypted web traffic classification under TLS 1.3~\cite{etbert2022}. We include it as an additional TLS 1.3 benchmark to test whether the learned concept space generalizes beyond CipherSpectrum to another modern encrypted traffic setting.

\noindent \textbf{ISCX-VPN(App) and ISCX-VPN(Service).}
These two benchmarks are derived from the ISCXVPN2016 dataset~\cite{DraperGil2016}. ISCX-VPN(App) uses application labels and evaluates application-level encrypted traffic classification, whereas ISCX-VPN(Service) uses coarser service-level labels. We include both settings to examine whether Traffic-CBM remains effective across related tasks with different label granularities.

\noindent \textbf{USTC-TFC2016.}
USTC-TFC2016 is a widely used benchmark in encrypted traffic analysis and malicious traffic classification~\cite{Wang2017MalwareCNN}. Compared with the application- and service-oriented benchmarks above, it provides a substantially different traffic classification scenario and thus serves as a complementary evaluation setting.

Overall, these six datasets provide complementary testbeds in terms of traffic source, protocol conditions, label granularity, and task type. Using them together allows us to assess not only predictive performance, but also whether the proposed hierarchical concept space remains effective across heterogeneous encrypted traffic benchmarks.

\section{Preprocessing Pipeline}

We use a two-stage preprocessing workflow to convert raw or pre-sliced traffic captures into aligned multimodal inputs for Traffic-CBM. The first stage prepares flow-level PCAP samples and split metadata, while the second stage constructs the statistical, sequential, and byte-level modalities used for downstream training.

In the first stage, all datasets are organized at the flow level. For datasets stored as raw packet captures, the original PCAP files are first converted into per-flow samples. For datasets that are already provided as flow-level PCAP files, the existing flow files are directly reused. In particular, CipherSpectrum is used as pre-sliced flow-level input after constructing an SNI-masked version of the dataset. We then apply filtering before feature construction. Flows with fewer than three packets are discarded, since they do not provide sufficient sequential or cross-packet context for the multimodal concept space. We also remove flows smaller than a dataset-specific minimum file-size threshold. In our experiments, this threshold is set to 512 bytes for USTC-TFC2016, CSTNET-TLS 1.3, and the two CipherSpectrum variants, and to 256 bytes for ISCX-VPN(App) and ISCX-VPN(Service). In the current setup, classes with fewer than 30 valid samples after filtering are excluded to avoid unstable train/validation/test splits. The remaining samples are divided using a class-wise stratified split with target ratios of 8:1:1. For small classes, lower-bound adjustment is applied when needed to ensure valid splits.

For CipherSpectrum, we additionally remove SNI-related information before feature construction. This step is necessary because early TLS handshake messages may expose service identifiers too strongly, thereby creating shortcut cues for classification~\cite{sok2025}. To address this issue, we construct an SNI-masked version of CipherSpectrum by masking only the \texttt{host\_name} field in the TLS ClientHello \texttt{server\_name} extension while preserving the original packet structure and TLS length fields. Concretely, the TCP payload stream is first reassembled for each direction, the TLS ClientHello is parsed, the \texttt{server\_name} extension is located, and only the corresponding \texttt{host\_name} bytes are overwritten. In our experiments, this masking is performed in zero mode. This yields a more behavior-oriented version of CipherSpectrum while minimizing disturbance to the remaining traffic structure.

To reduce shortcut cues more generally, we avoid directly using explicit connection identifiers as model inputs~\cite{sok2025}. In particular, the byte-level extraction starts from a fixed offset rather than from the beginning of each packet, which reduces direct exposure to low-level identifier fields and initial handshake artifacts~\cite{sok2025}. For CipherSpectrum, this is further combined with the SNI-masked construction described above.

In the second stage, we construct three modalities from the filtered flow-level samples. The statistical modality captures flow-level macro properties using CICFlowMeter-style flow statistics, including packet counts, packet-length statistics, timing statistics, flow bytes, TCP flags, and header-related information~\cite{cicflowmeter}. In the current setup, statistical features are extracted using up to 200 packets per flow. The sequential modality captures packet-level behavioral dynamics by representing each flow as an ordered packet sequence, where each packet is described by seven attributes: packet length, inter-arrival time, direction, TCP flags, payload length, header length, and TCP window size. In the current setup, the sequential branch uses up to 100 packets per flow. The byte modality captures fine-grained packet content by extracting a fixed 64-byte window from each of the first five packets, starting from offset 38, and converting each window into a sequence of bigram tokens. This packet-wise byte construction is broadly consistent with the fixed-packet, fixed-byte input paradigm adopted in recent pre-training-based traffic classifiers~\cite{etbert2022,trafficformer2025}. In the final model, these packet-wise token sequences are used to form both packet-level and cross-packet byte concepts. Together, the three modalities provide flow-level statistical evidence, packet-level temporal evidence, and byte-level structural evidence for Traffic-CBM.

After all three modalities are constructed, we perform a final alignment step and retain only flows for which all three modalities are available. In other words, downstream training, validation, and testing are conducted on the intersection of the statistical, sequential, and byte-level samples. This alignment procedure ensures that every sample used by the model has a complete multimodal representation, avoids inconsistencies across modalities, and keeps label indexing and data loading reproducible across all experiments.

\section{Detailed Model and Training Configuration}

\subsection{Model Configuration}

Unless otherwise stated, all experiments use the full concept configuration of Traffic-CBM, yielding an 18-dimensional concept representation composed of 6 statistical concepts, 4 temporal concepts, 5 packet-level byte concepts, and 3 cross-packet byte concepts.

\noindent \textbf{Statistical branch.}
The statistical branch uses the grouped concept setting, where flow-level statistical features extracted from CICFlowMeter-style statistics~\cite{cicflowmeter} are partitioned into six predefined groups corresponding to packet count, packet length, flow bytes, packet time, flag count, and header/window information. These groups are implemented as predefined feature-name sets, and each group is mapped to one scalar concept by an independent multilayer perceptron with hidden dimensions \([128,128,128,64]\).

\noindent \textbf{Sequential branch.}
The sequential branch operates on the first 100 packets of each flow. Each packet is represented by seven attributes, namely packet length, inter-arrival time, direction, TCP flags, payload length, header length, and TCP window size, following a local packet-feature formulation in line with recent encrypted traffic classifiers~\cite{swift2025}. The sequential concept encoder contains four physically separated Transformer encoders~\cite{vaswani2017transformer} corresponding to the \emph{volume}, \emph{timing}, \emph{direction}, and \emph{TCP} groups. Concretely, the volume group uses packet length, payload length, and header length; the timing group uses inter-arrival time; the direction group uses packet direction; and the TCP group uses TCP flags and TCP window size. Each encoder uses model dimension 64, 4 attention heads, 2 Transformer layers, and \texttt{CLS} pooling.

\noindent \textbf{Byte branch.}
The byte branch uses the first five packets of each flow. Following packet-wise byte input formulations commonly adopted in recent pre-training-based traffic classifiers~\cite{etbert2022,trafficformer2025}, we extract a 64-byte window from each packet starting from offset 38 and convert it into 63 bigram tokens. The resulting byte input has shape \(B \times 5 \times 63\). The packet-level pathway produces five scalar concepts, one for each packet. The cross-packet pathway further divides each packet-wise token sequence into three aligned regions and aggregates the corresponding regions across the five packets to produce three additional scalar concepts. In our implementation, each packet sequence of length 63 is evenly divided into three segments of length 21, corresponding to the front, middle, and back regions. Both the packet-level and cross-packet byte encoders use model dimension 64, 4 attention heads, 2 Transformer layers~\cite{vaswani2017transformer}, and \texttt{CLS} pooling.

Within the byte branch, the five packet-level concepts share one packet-wise Transformer encoder, while the three cross-packet concepts share a separate cross-packet Transformer encoder. The packet-level and cross-packet pathways do not share Transformer parameters, positional embeddings, or \texttt{CLS} tokens, although they share the same token embedding layer.

\noindent \textbf{Prediction head.}
The final prediction is performed on the 18-dimensional concept vector using a GA2M head~\cite{Lou2013GA2M,caruana2015ga2m}. Unless otherwise stated, the model uses dropout 0.2. The main paper reports results from this GA2M-based setting because it supports both first-order and pairwise concept-level analysis.

Table~\ref{tab:hyperparams} summarizes the main hyperparameters used in the final model and training setup, and Table~\ref{tab:concept_groups} summarizes the concept grouping used in the statistical and sequential branches.

\begin{table}[t]
\centering
\caption{Main hyperparameters used in Traffic-CBM.}
\vspace{-5pt}
\label{tab:hyperparams}
\scriptsize
\setlength{\tabcolsep}{4pt}
\begin{tabular}{ll}
\toprule
\textbf{Item} & \textbf{Setting} \\
\midrule
Concepts (stat / temp / pkt-byte / cross-byte) & 6 / 4 / 5 / 3 \\
Stat branch hidden dims & {[}128, 128, 128, 64{]} \\
Stats max packets & 200 \\
Seq max packets & 100 \\
Seq features per packet & 7 \\
Byte packets used & 5 \\
Byte window length & 64 bytes \\
Byte start offset & 38 \\
Byte tokens per packet & 63 bigram tokens \\
Encoder type & Transformer \\
Model dimension & 64 \\
Attention heads & 4 \\
Transformer layers & 2 \\
Pooling & CLS \\
Prediction head & GA2M \\
Dropout & 0.2 \\
Optimizer & AdamW \\
Batch size & 256 \\
Learning rate & \(3\times 10^{-4}\) \\
Weight decay & \(10^{-2}\) \\
Label smoothing & 0.05 \\
Max epochs & 80 \\
Early stopping patience & 10 \\
Hardware & NVIDIA RTX 4090 \\
\bottomrule
\end{tabular}
\vspace{-8pt}
\end{table}

\begin{table*}[t]
\centering
\caption{Concept grouping used in the statistical and sequential branches.}
\vspace{-4pt}
\label{tab:concept_groups}
\scriptsize
\setlength{\tabcolsep}{6pt}
\begin{tabular}{lll}
\toprule
\textbf{Branch} & \textbf{Group} & \textbf{Representative Features} \\
\midrule
Stat & packet count &
\texttt{tot\_fwd\_pkts}, \texttt{tot\_bwd\_pkts}, \texttt{fwd\_pkts\_s}, \texttt{bwd\_pkts\_s} \\
Stat & packet length &
\texttt{fwd\_pkt\_len\_mean}, \texttt{bwd\_pkt\_len\_mean}, \texttt{pkt\_len\_mean}, \texttt{pkt\_len\_std} \\
Stat & flow bytes &
\texttt{flow\_byts\_s}, \texttt{flow\_pkts\_s}, \texttt{down\_up\_ratio}, \texttt{subflow\_fwd\_byts} \\
Stat & packet time &
\texttt{flow\_duration}, \texttt{flow\_iat\_mean}, \texttt{fwd\_iat\_mean}, \texttt{active\_mean}, \texttt{idle\_mean} \\
Stat & flag count &
\texttt{syn\_flag\_cnt}, \texttt{ack\_flag\_cnt}, \texttt{psh\_flag\_cnt}, \texttt{rst\_flag\_cnt} \\
Stat & header/window &
\texttt{fwd\_header\_len}, \texttt{bwd\_header\_len}, \texttt{init\_fwd\_win\_byts}, \texttt{init\_bwd\_win\_byts} \\
\midrule
Seq & volume &
\texttt{pktlen}, \texttt{payload\_len}, \texttt{header\_len} \\
Seq & timing &
\texttt{iat} \\
Seq & direction &
\texttt{direction} \\
Seq & tcp &
\texttt{tcp\_flags}, \texttt{tcp\_window} \\
\bottomrule
\end{tabular}
\vspace{-4pt}
\end{table*}

\begin{table*}[!t]
\centering
\caption{Efficiency comparison of Traffic-CBM and representative baselines.}
\vspace{-5pt}
\label{tab:efficiency}
\scriptsize
\setlength{\tabcolsep}{5pt}
\begin{tabular}{lccccc}
\toprule
\textbf{Method} & \textbf{Params (M)} & \textbf{FLOPs (G)} & \textbf{Latency (ms/sample)} & \textbf{Throughput (samples/s)} & \textbf{GPU Mem (MB)} \\
\midrule
AppScanner~\cite{appscanner2018} & -- & -- & 0.240 & 4167.0 & -- \\
DeepFP~\cite{deepfp2018} & 2.1 & -- & 0.621 & 1609.7 & -- \\
FS-Net~\cite{fsnet2019} & 1.6 & -- & 100.986 & 9.9 & -- \\
ET-BERT~\cite{etbert2022} & 102.3 & 715.8 & 0.865 & 1155.6 & 660.6 \\
YaTC~\cite{yatc2023} & 1.9 & 57.9 & 0.345 & 2902.0 & 69.9 \\
TrafficFormer~\cite{trafficformer2025} & 102.3 & 357.6 & 0.929 & 1076.3 & 961.6 \\
NetMamba~\cite{netmamba2024} & 1.9 & 43.7 & 0.084 & 11836.7 & 324.8 \\
\midrule
\textbf{Traffic-CBM} & \textbf{5.2} & \textbf{5.9} & \textbf{3.367} & \textbf{297.0} & \textbf{64.4} \\
\bottomrule
\end{tabular}
\end{table*}

\subsection{Training Configuration}

All models are trained with AdamW~\cite{adamw2019} for at most 80 epochs using batch size 256, initial learning rate \(3\times10^{-4}\), and weight decay \(10^{-2}\). We apply cosine annealing over training epochs. Label smoothing is set to 0.05. Model selection is based on validation loss, and early stopping is applied with patience 10. After training, the checkpoint with the best validation loss is used for final test evaluation.

For the main experiments, all model parameters, including the statistical, sequential, byte, and prediction-head modules, are optimized jointly in a single supervised stage. Although the implementation supports branch-specific learning rates, in the main setting we use the same learning rate for all trainable modules unless otherwise specified in ablation experiments.

\noindent \textbf{Self-supervised initialization.}
Before supervised training, the sequential and byte encoders are optionally initialized from a cross-dataset self-supervised pretraining stage. In this stage, the sequential branch is trained with masked reconstruction of packet-level sequential features, analogous in spirit to masked autoencoding~\cite{mae2022}, while the byte branch is trained with masked token reconstruction on bigram token sequences, analogous to masked language modeling~\cite{bert2019}. The pretrained checkpoint is then loaded to initialize the sequential and byte encoders in downstream supervised training. We treat this initialization as an auxiliary strategy rather than a core component of the method, and its contribution is analyzed separately in the ablation study.

\noindent \textbf{Implementation environment.}
All experiments are conducted on a server equipped with an NVIDIA RTX 4090 GPU. All compared methods are trained and evaluated on the same processed split. When reproducing baselines, we follow their original settings as closely as possible while keeping the evaluation protocol consistent.

\section{Efficiency Analysis}

Table~\ref{tab:efficiency} reports the efficiency comparison of Traffic-CBM and representative baselines under their native input formats and inference settings. Following common practice in implementation-oriented comparison, we report parameter count, estimated FLOPs when reliable estimation is available, inference latency, throughput, and GPU memory usage. Since the compared methods are implemented under different input paradigms and software stacks, these results should be understood as a practical efficiency reference rather than a perfectly unified hardware-level comparison. For methods implemented outside the current PyTorch measurement pipeline, some efficiency indicators such as FLOPs or GPU memory are omitted when reliable estimation is unavailable.

From the results, Traffic-CBM uses a moderate number of parameters and relatively low GPU memory compared with large pre-trained Transformer baselines such as ET-BERT and TrafficFormer. In particular, although Traffic-CBM is slower in inference than several highly optimized black-box baselines, its memory footprint remains comparatively small. This is consistent with the design of the model: Traffic-CBM introduces structured multimodal concept branches and a concept-level prediction head, which improve analyzability but also add extra computation in the forward pass.

The results also illustrate a broader trade-off between efficiency and model design. Methods such as NetMamba and YaTC achieve substantially lower latency and higher throughput, whereas Traffic-CBM emphasizes structural interpretability and concept-level analysis rather than raw inference efficiency alone. Therefore, the efficiency results should be interpreted together with the main performance and interpretability analyses in the paper.

\section{Limitations and Discussion}

Although Traffic-CBM provides a more structured interpretation interface than conventional end-to-end multimodal classifiers, its interpretability remains partial rather than complete. In particular, the statistical and temporal concepts have relatively clear semantic boundaries because they are tied to predefined feature groups or physically separated sequential subspaces. By contrast, the byte-level concepts are primarily structural concepts. They can indicate which packet-level or cross-packet byte patterns are important, but they do not always admit a direct protocol-semantic interpretation in terms of specific handshake fields or cryptographic primitives.

Moreover, byte-level analysis is still mainly conducted at the concept level rather than at the token or position level. While the learned byte concepts reveal which packet-level or cross-packet patterns are important, they do not yet localize the most influential byte regions within each concept pathway. A possible extension is to incorporate finer-grained attention-based analysis in the byte branch, which may help identify salient packet regions or cross-packet correspondences and provide a more detailed interpretation of byte-level evidence.

The full hierarchical concept configuration used in this work should also be understood as a design trade-off rather than a universally optimal decomposition. It provides a practical balance between predictive competitiveness and structural interpretability: simpler variants remove part of the byte-level structure, while more performance-oriented variants may provide a less direct interpretation interface. Accordingly, we view the present configuration as a useful analysis setting rather than as the only valid concept decomposition for encrypted traffic classification.

Finally, despite the preprocessing steps used to reduce shortcut cues, some residual dataset-specific bias may still remain. Although we avoid directly feeding explicit connection identifiers into the model and additionally construct an SNI-masked version of CipherSpectrum, indirect statistical regularities or collection-specific artifacts may still influence model behavior. More systematic evaluation under cross-collection or cross-environment transfer would therefore be valuable for assessing the robustness and generalizability of the learned concept space.

\end{document}